\documentclass[acmtog]{acmart}
\AtBeginDocument{%
  }

\setcopyright{acmcopyright}
\copyrightyear{2018}
\acmYear{2018}
\acmDOI{XXXXXXX.XXXXXXX}

\acmJournal{TOG}
\acmVolume{37}
\acmNumber{4}
\acmArticle{111}
\acmMonth{8}

\begin{document}

\title{High-Resolution Volumetric Reconstruction for Clothed Humans}

\author{Sicong Tang}
\authornote{Both authors contributed equally to this research.}
\email{sta105@sfu.ca}
\affiliation{%
  \institution{Simon Fraser University}
  \city{Vancouver}
  \country{Canada}
}

\author{Guangyuan Wang}
\authornotemark[1]
\email{yixuan.wgy@alibaba-inc.com}
\affiliation{%
  \institution{Alibaba Group}
  \city{Hang Zhou}
  \country{China}
}
\author{Qing Ran}
\email{ranqing.rq@alibaba-inc.com}
\affiliation{%
  \institution{Alibaba Group}
  \city{Hang Zhou}
  \country{China}
}
\author{Lingzhi Li}
\email{llz273714@alibaba-inc.com}
\affiliation{%
  \institution{Alibaba Group}
  \city{Hang Zhou}
  \country{China}
}
\author{Li Shen}
\email{lshen.lsh@gmail.com}
\affiliation{%
  \institution{Alibaba Group}
  \city{Hang Zhou}
  \country{China}
}

\author{Ping Tan}
\affiliation{%
  \institution{Simon Fraser University}
  \city{Vancouver}
  \country{Canada}}
\email{pingtan@sfu.ca}

\renewcommand{\shortauthors}{Tang et al.}

\begin{abstract}
We present a novel method for reconstructing clothed humans from a sparse set of, e.g., 1--6 RGB images. Despite impressive results from recent works employing deep implicit representation, we revisit the volumetric approach and demonstrate that better performance can be achieved with proper system design. The volumetric representation offers significant advantages in leveraging 3D spatial context through 3D convolutions, and the notorious quantization error is largely negligible with a reasonably large yet affordable volume resolution, e.g., 512. To handle memory and computation costs, we propose a sophisticated coarse-to-fine strategy with voxel culling and subspace sparse convolution. Our method starts with a discretized visual hull to compute a coarse shape and then focuses on a narrow band nearby the coarse shape for refinement. Once the shape is reconstructed, we adopt an image-based rendering approach, which computes the colors of surface points by blending input images with learned weights. Extensive experimental results show that our method significantly reduces the mean point-to-surface (P2S) precision of state-of-the-art methods by more than 50\% to achieve approximately 2mm accuracy with a 512 volume resolution. Additionally, images rendered from our textured model achieve a higher peak signal-to-noise ratio (PSNR) compared to state-of-the-art methods.

\end{abstract}



\received{20 February 2007}
\received[revised]{12 March 2009}
\received[accepted]{5 June 2009}

\maketitle
\begin{figure}
\begin{center}
\includegraphics[width=\linewidth]{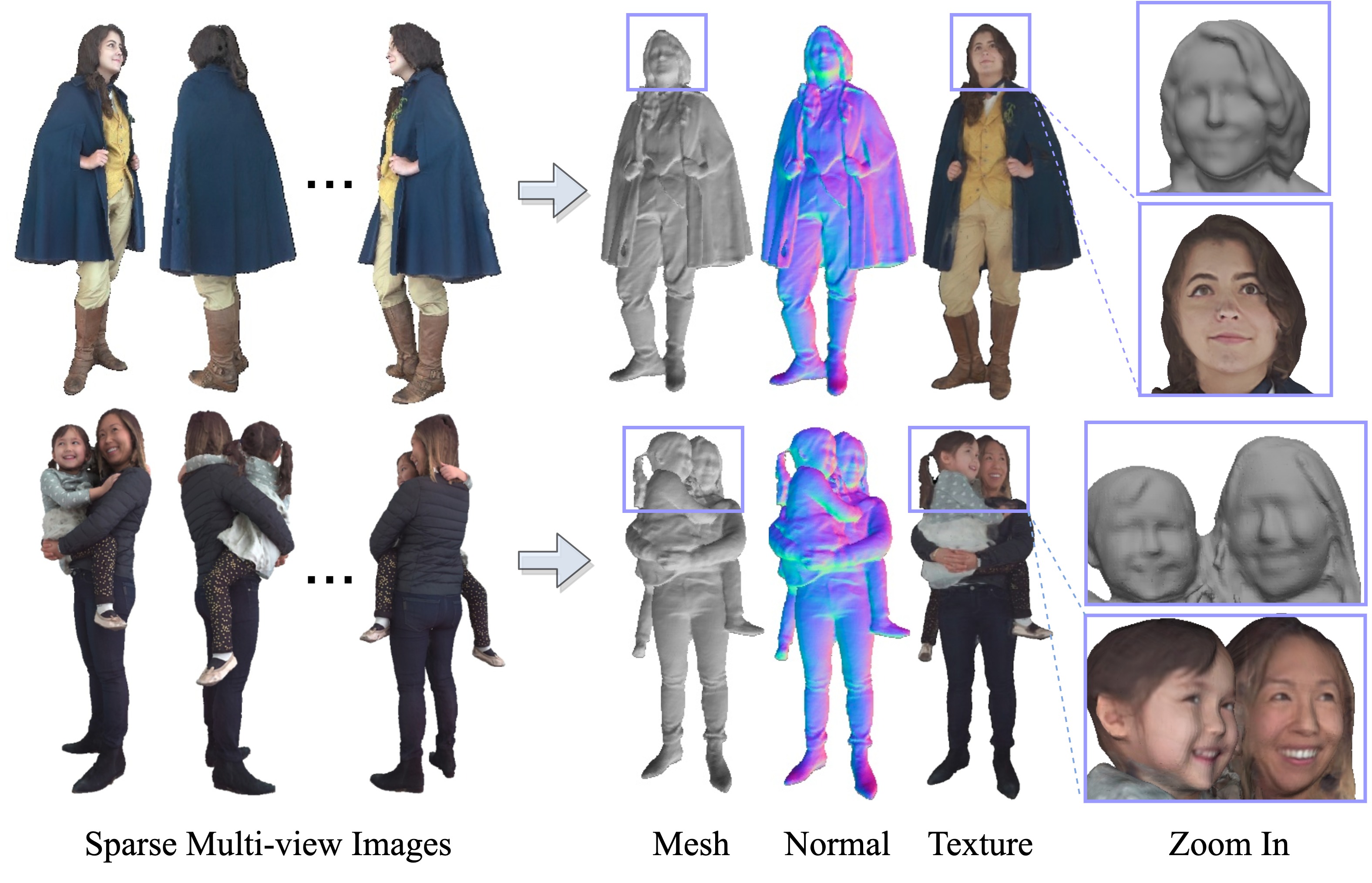}
\end{center}
\vspace{-0.15in}
    \caption{Our method reconstructs a textured 3D model of clothed humans from sparse multi-view images. It recovers detailed geometry with vivid texture, despite complexities caused by garments, poses, and occlusions.}
\label{fig:teaser}
\vspace{-0.15in}
\end{figure}

\section{Introduction}
Automatic 3D reconstruction of clothed humans using image inputs has gained increasing significance due to its potential applications in a wide array of AR/VR scenarios. High-fidelity reconstructions typically depend on sophisticated capture systems, which are developed with dense camera arrays~\cite{collet2015high,joo2015panoptic,joo2018total}, programmable light-stages~\cite{Vlasic2009, guo2019relightables}, and depth sensors~\cite{newcombe2011kinectfusion,DoubleFusion,BodyFusion,dou2016fusion4d,newcombe2015dynamicfusion}. However, stringent capture environments equipped with complex hardware pose significant challenges for consumer-level applications.

In this context, considerable research effort has been dedicated to developing methods that allow for more flexible capture configurations, such as utilizing a few RGB inputs. Among these works, learning implicit functions \cite{iccv2020PIFu, saito2020pifuhd, hong2021stereopifu} has proven effective in achieving highly detailed reconstructions by integrating the advancements of deep neural networks. These methods employ large multi-layer perceptrons (MLPs) to predict the occupancy probability or truncated signed distance function (TSDF) value of every queried 3D point based on its associated local feature, which is extracted from images. They can recover a continuous surface at arbitrary resolutions without topology restrictions.

However, in typical MLP-based implicit networks, the occupancy or TSDF value at each location is solved independently with planar image features, rendering them less capable of addressing challenging cases such as occlusions. Consequently, these methods suffer from generalization and robustness issues, particularly when tackling strong occlusions caused by large motion or multiple interacting humans. 
Some follow-up studies  \cite{zheng2021deepmulticap,zheng2021pamir,huang2020arch} utilize an extra geometric model, SMPL~\cite{Loper2015}, to improve robustness by introducing strong shape priors. 
Their success typically relies on the assumption of geometrical similarity \cite{huang2020arch} between the shape prior and target reconstruction, making them intractable for handling complex cases with loose clothes and sensitive to errors in SMPL model fitting.



\begin{figure}
\begin{center}
    \includegraphics[width=1.0\linewidth]{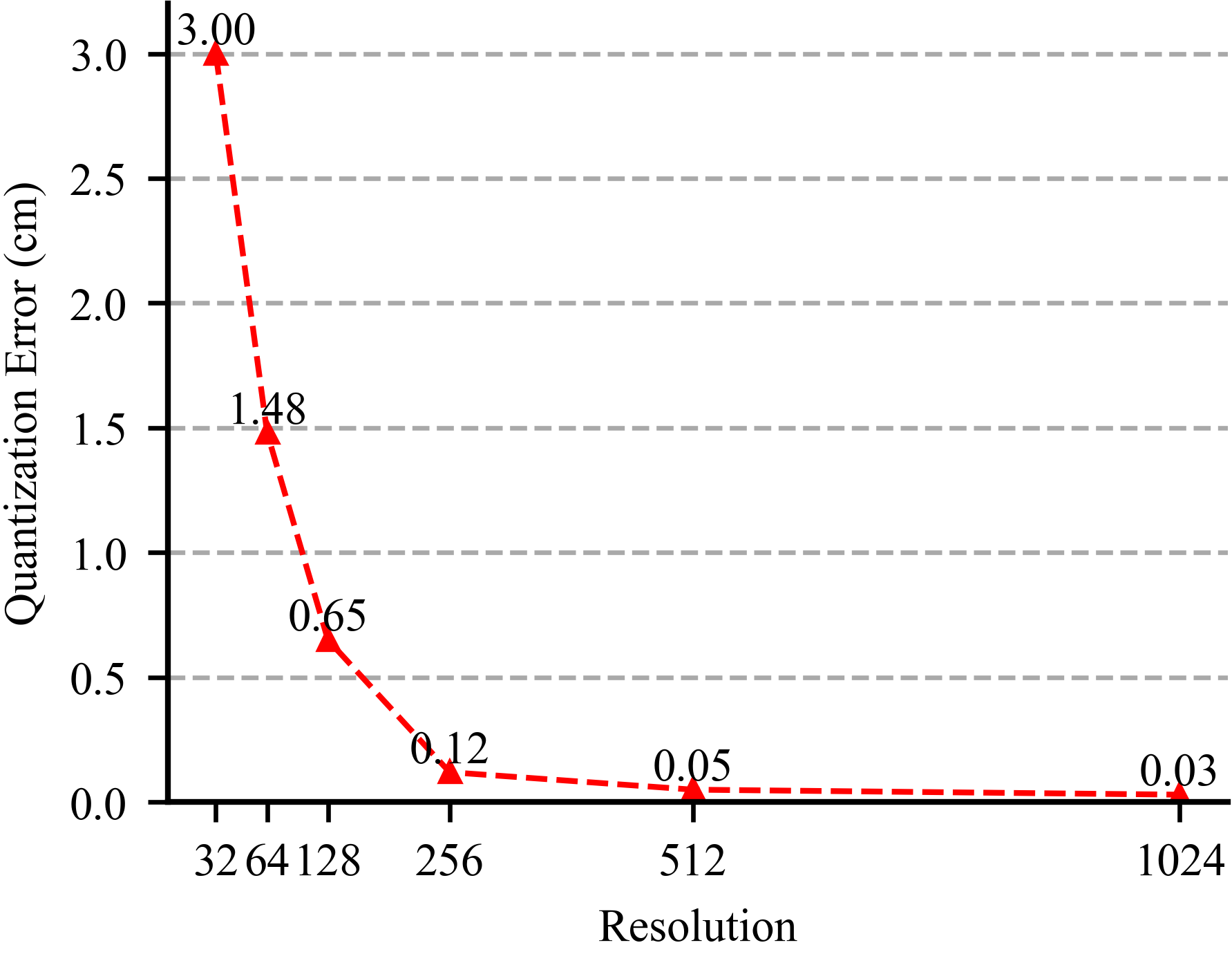}
\end{center}
\vspace{-0.15in}
    \caption{Quantization error of TSDF volume at different resolutions. The quantization error becomes negligible  for volume resolutions beyond 512.
    }
\label{fig:quantization_error}
\vspace{-0.1in}
\end{figure}

We instead revisit the 3D volumetric representation and resort to 3D convolutional neural networks (CNNs) for feature learning, due to their impressive performance in feature learning and the ability to incorporate spatial context. However, volumetric methods and 3D convolution involve discretization, which might raise concerns regarding whether a discretized volume can preserve subtle geometric details as continuous representations learned in implicit functions. We investigate the relationship between volume resolution and quantization error on synthetic data by converting target mesh objects to TSDF volumes, as shown in Figure~\ref{fig:quantization_error}. We observe that the quantization errors are significantly reduced by increasing volume resolution and become nearly negligible when reaching a relatively high resolution (e.g., 512 or higher). In other words, achieving fine-detailed reconstruction is not supposed to be restricted by the use of volume representations as long as a proper volume resolution is utilized. Therefore, we present a method with high-resolution feature volumes, e.g., 256 and 512, while traditional volumetric methods \cite{varol18_bodynet,gilbert2018volumetric} are often limited to much lower resolutions, such as 32 or 128.

On the other hand, an increase in volume resolution may lead to a cubic growth of memory overhead \cite{8100085}. Reducing memory costs while guaranteeing the granularity of volumetric representations is necessary for pursuing high-quality reconstruction. Thus, we adopt a coarse-to-fine approach and cull away irrelevant voxels to build a sparse high-resolution feature volume. At the coarse level, the network computes an initial TSDF by applying a U-Net with sparse 3D CNN \cite{3DSemanticSegmentationWithSubmanifoldSparseConvNet} on the sparse feature volume, which is carved by a visual hull. Through our experiments, it turns out that more than 95\% of the volume grids are discarded by the visual hull culling, making the sparse 3D CNN efficient. At the fine level, the network focuses on a narrow band near the zero-level set of the initial TSDF and discretizes the narrow band with smaller voxels. By employing this narrow-band culling, we further shrink the sampling space, resulting in a relatively small range of grid numbers (usually 300K--500K in our experiments) even with a high volume resolution of 512. The remaining voxels in the narrow band are associated with features that fuse high-frequency information from the computed normal maps upon the low-frequency shape from the coarse level to compute the TSDF at high resolution. The final mesh is then extracted from the TSDF using the Marching-Cube algorithm ~\cite{Lorensen87marchingcubes}.


In addition to geometry, high-quality mesh texture is also a crucial factor contributing to visual appearance. Directly computing a color field in 3D space, as in \cite{iccv2020PIFu}, struggles to capture high-frequency texture details, while the neural radiance field (NeRF) \cite{yu2020pixelnerf} or the DoubleField~\cite{shao2022doublefield} require expensive per-instance optimization and are often unstable for sparse input images. In contrast, we adopt an image-based rendering approach to compute a texture atlas map, which is efficient and widely supported in existing computer graphics tools. 
Specifically, we compute a blending weight at each 3D point on the mesh surface to determine its color as a weighted average of the colors at its image projections. The blending weights can be computed at a relatively coarse resolution, e.g., 512 volume resolution in our case, and leave texture details to the high-resolution images, such as 1K or 2K. Unlike previous methods that generate blurry texturing results under sparse input, our method generalizes well on both synthetic and real data with just a few input views. 
Figure~\ref{fig:teaser} shows two examples reconstructed by our method. Despite the challenging garment, pose, and occlusion, our method recovers faithful shape, normal, and texture on the right.


In summary, the main contributions of this paper are as follows:
\begin{itemize}
\vspace{-0.1in}
  \item 
  We revisit the 3D volumetric representation and demonstrate that it can support clothed human reconstruction with equal or even better performance compared to implicit representation. 
  \item 
  We develop a memory and computation-efficient method for high-resolution volumetric reconstruction using sophisticated sparse 3D CNN, coarse-to-fine estimation, and voxel culling by visual hull and narrow bands. 
  \item 
  We introduce a novel method to compute a texture atlas map, which captures rich appearance details from high-resolution input images.
  \item 
  We achieve impressive results on standard benchmark datasets Twindom and MultiHuman, significantly reducing the point-2-surface (P2S) precision to approximately 0.2cm from just six input views, with more than $50\%$ error reduction compared to the state-of-the-art methods, including DoubleField~\cite{shao2022doublefield} and PIFuHD~\cite{saito2020pifuhd}.
\end{itemize}
\section{Related work}
\noindent\textbf{Parametric Model Based Methods.}
Parametric human models such as SCAPE~\cite{anguelov2005scape}, SMPL~\cite{Loper2015}, and SMPL-X~\cite{pavlakos2019expressive} have been widely adopted to recover human pose and shapes. SMPLify~\cite{bogo2016keep}  estimates the SMPL model from a single image using 2D keypoint detection. 
Recently, deep neural networks~\cite{hmrKanazawa17,omran2018neural,pavlakos2018learning, xu2019denserac} have been trained to directly regress SMPL model parameters from a single image. The accuracy is further improved by combining bottom-up optimization~\cite{guler2019holopose, kolotouros2019learning} and using temporal generation networks~\cite{kocabas2020vibe}.
However, the SMPL estimation from a single image suffers from shape ambiguity. Huang et al.~\shortcite{3dv2017Towards} and Liang et.al~\shortcite{liang2019shapeaware} generalize SMPL model fitting to multiple input images. 
To capture cloth shape details, methods like ~\cite{alldieck2019tex2shape,bhatnagar2019mgn} use SMPL+D representation to explain high-frequency details. However, this representation struggles to handle loose clothes and long hair. In contrast to  these methods, we aim to reconstruct a clothed human without relying on any parametric model to achieve better generalization to different poses and garments.

\noindent\textbf{Implicit Function Based Methods.} 
Implicit functions~\cite{Mescheder2019,Park2019,Chen2019} provide a powerful shape representation for 3D reconstruction, enabling surface reconstruction at arbitrary resolutions and topologies.
Many recent works utilize implicit functions to reconstruct clothed humans. Huang et al.~\shortcite{Huang18ECCV} and PIFu~\cite{iccv2020PIFu} introduce implicit functions for clothed human reconstruction, where a 3D point is projected onto the input images to gather features for occupancy regression. PIFuHD~\cite{saito2020pifuhd} improves PIFu by extracting high-resolution features from estimated normal maps. PHORHUM~\cite{alldieck2022phorhum} also predicts shading parameters to generate  more realistic rendering results. MonoPort~\cite{li2020monoport} accelerates the occupancy evaluation in a coarse-to-fine manner using Octree structures.

However, the aforementioned methods compute the occupancy or TSDF of a 3D point using point-wise inference and planar image features, which are limited in exploring 3D context information, especially for self-occlusions and challenging poses.
As a result, Arch~\cite{huang2020arch}, Arch++~\cite{He2021ARCHAC}, PaMIR \cite{zheng2021pamir}, and ICON~\cite{xiu2022icon} employ the SMPL model as a shape prior to improve robustness to different body poses. DeepMultiCap~\cite{zheng2021deepmulticap} further employs a spatial attention network and a temporal fusion method for multi-view input videos. 
However, the SMPL model estimation itself is fragile, StereoPIFu \cite{hong2021stereopifu} takes stereo images as input to exploit the geometric constraints of stereo vision. 
The recent work, DoubleField~\cite{shao2022doublefield}, combines the neural radiance field with an implicit surface field to  generate high-quality results. DiffuStereo~\cite{shao2022diffustereo} further introduces a diffusion-based stereo algorithm to enhance shape accuracy by enforcing multi-view correspondences. 

As observed in \cite{chibane20ifnet,Peng2020ECCV},  implicit function based 3D reconstruction can benefit from a 3D convolutional feature encoder. While these two methods are designed for 3D reconstruction from point clouds or sparse voxel inputs, we extend a similar idea for image-based reconstruction of clothed humans. 3D convolutions can easily encode geometric contexts and compute the TSDF values at nearby points jointly. However, it requires a high-resolution feature volume to capture shape details.




\noindent\textbf{Volumetric Methods.} 
There are relatively fewer works adopting volumetric representation for clothed human reconstruction, as it is known to have expensive memory and running time costs. 
To reduce memory and computation costs, earlier works~\cite{varol18_bodynet,Jackson20183DHB} employ 2D convolutions to regress the occupancy volume from a single RGB image, but only recover limited shape details and suffer from challenging poses. DeepHuman~\cite{Zheng2019DeepHuman} employs 3D convolution and uses SMPL models as shape priors to guide the volume regression. However, like parametric model-based methods, its SMPL estimation tends to fail at challenging poses and loose garments. Gilbert et al.~\shortcite{gilbert2018volumetric} use multi-view images to recover a visual hull, and then apply 3D CNN to compute the occupancy values at the discretized visual hull. However, they do not involve any image features in the 3D CNN, which is crucial to recover shape details, and only generate over-smoothed results. Similar 3D convolution is also applied to the discretized SMPL model in \cite{zheng2021pamir, zheng2021deepmulticap} to facilitate learning implicit functions, but image features are not involved in the 3D convolution. 
Furthermore, most previous methods~\cite{varol18_bodynet,Jackson20183DHB,Zheng2019DeepHuman,zheng2021pamir,zheng2021deepmulticap} use low volume resolution of 128, except the method in \cite{gilbert2018volumetric} uses 256 volume resolution without including image features. 

We find it is important to use high-resolution volumes, e.g. 512 or higher, for accurate 3D reconstruction of clothed humans. Furthermore, it is important to include image features in the 3D convolution to reconstruct shape details. To make these ideas feasible, we design a sophisticated volumetric method by combining efficient sparse 3D convolution, voxel culling, and coarse-to-fine computation. 



\noindent\textbf{Surface Color Prediction.} 
Traditional methods like \cite{Waechter2014Texturing} color surface points according to images with front parallel viewing directions. For human modeling tasks, PIFu~\cite{iccv2020PIFu} and its follow-up works \cite{zheng2021deepmulticap,tao2021function4d} often use an additional implicit function to compute a continuous color field, where each 3D point is associated with a color. Implicit functions can hallucinate colors in unobserved regions. However, it is also difficult to capture appearance details like high-frequency textures due to the compact representation.
Recently, neural radiance field (NeRF) ~\cite{mildenhall2020nerf} has shown its great potential of generating high quality view synthesis.
NeuralBody ~\cite{peng2021neural} further introduces a SMPL model to aggregate color constraints in the canonical frame, extending the NeRF-based human reconstruction to sparse multi-view inputs. 
To capture 3D shape details while generating realistic rendering, Doublefield ~\cite{shao2022doublefield} combines the advantages of implicit surface field~\cite{iccv2020PIFu} and neural radiance field ~\cite{mildenhall2020nerf}, which further speedup the convergence of NeRF models.
While these methods generate high-quality results, NeRF-based methods still require expensive per-instance optimization, which is undesirable in many real applications.
To address these problems, we follow the spirit of traditional methods to compute a texture map on the mesh surface. Instead of computing the texture color directly, we design a network to estimate a blending weight to evaluate the color according to the input images. In this way, our texture map can easily inherit high-frequency details from high-resolution input images.

\begin{figure*}
\begin{center}
    \includegraphics[width=1.0\linewidth]{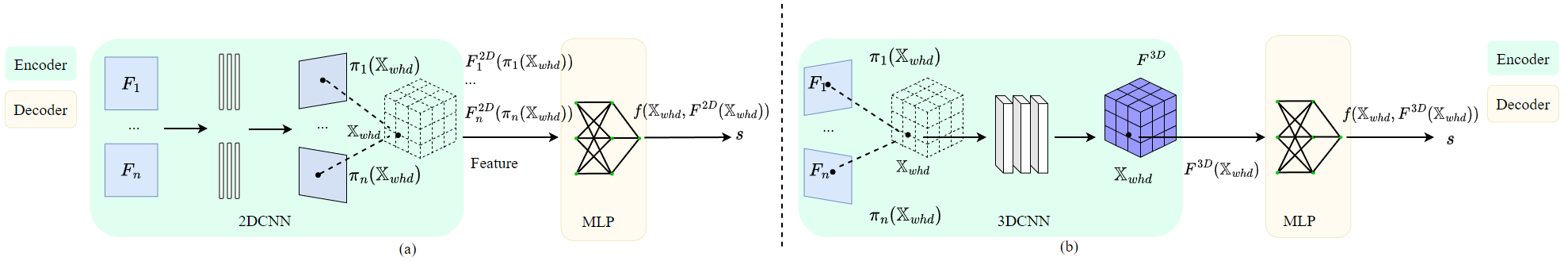}
\end{center}
\vspace{-0.15in}
    \caption{The network architecture of the two toy networks in Section~\ref{sec:feature}. (a) features are encoded in the 2D image plane, similar to the multi-plane encoder proposed in \cite{Peng2020ECCV}.
    (b) features are encoded in the 3D volume.}
\label{fig:toy_ablation}
\vspace{-0.15in}
\end{figure*}

\section{Motivation}
\label{sec:motivation}
Given sparse view RGB images $\{ \mathbf{I}_i \}$ capturing a clothed human and their calibrations, our goal is to estimate the truncated signed distance function (TSDF) $\mathbb{D}$ which describes the clothed human shape. This TSDF $\mathbb{D}$ might be directly discretized as a 3D volume, where each voxel grid stores the TSDF value. In contrast, recent learning based reconstruction methods~\citep{Mescheder2019,Park2019,Chen2019} employ a  multi-layer perceptron (MLP) as an implicit representation of the TSDF or occupancy function, which is continuous and free from resolution and topology limitations. Following this idea, PIFu\cite{iccv2020PIFu} computes the occupancy function with pixel-aligned features as,
\begin{equation}
	\mathbb{D} (\mathbf{X}) = f (\mathbf{X}, \mathbf{F}^{2D}(\Pi( \mathbf{X}))) = s, \qquad s\in [-1,1],
	\label{equation:2Dfeature}
\end{equation}
where $f$ is an MLP, $\mathbf{X}$ is a 3D point, and $\Pi(\cdot)$ projects 3D points to the input image. The feature map $\mathbf{F}^{2D}$ is computed from the input image with 2D convolutions.

On the other hand, the simple fully-connected network architecture of MLPs is inefficient in integrating context information as studied in \cite{chibane20ifnet,Peng2020ECCV}. These studies suggest combining a 3D convolutional encoder with an MLP decoder for 3D reconstruction from point clouds. In the same spirit, we might solve the TSDF $\mathbb{D}$ as
\begin{equation}
	\mathbb{D} (\mathbf{X}) = f (\mathbf{X}, \mathbf{F}^{3D}( \mathbf{X})) = s, \qquad s \in [-1,1],
	\label{equation:3Dfeature}
\end{equation}
where the feature volume $\mathbf{F}^{3D}$ is computed in 3D by a 3D convolutional encoder. As reported in ~\cite{Peng2020ECCV}, learning this 3D feature volume is superior to its counterpart 2D feature maps (i.e. the single-plane or multi-plane feature encoder), which are commonly employed in earlier clothed human reconstruction methods including PIFu\cite{iccv2020PIFu}, PaMIR\cite{zheng2021pamir}, and DeepMultiCap\cite{zheng2021deepmulticap}.



\begin{table}
\begin{tabular}{|l|l|l|l|l|l|}
\hline
 & 2D Features  & 3D Features   \\ \hline
 Chamfer/P2S&  0.601/0.549 &  0.404/0.358    \\ \hline
\end{tabular}
\caption{Chamfer and P2S precision errors of the two toy networks with 2D or 3D features tested on the Twindom\cite{twindom} dataset. }
\label{tab:ablation_palvo}
\vspace{-0.1in}
\end{table}

\subsection{3D Feature Volume}\label{sec:feature}
To demonstrate the strength of 3D convolution in Equation~\ref{equation:3Dfeature}, we experiment with a toy network by directly appending a 3D convolutional feature encoder with an MLP decoder for clothed human reconstruction. 
Specifically, as in Figure~\ref{fig:toy_ablation} (b), we sample a set of regular volume grid $\{ \mathbb{X}_{whd} \}$ in 3D space, where $\{w, h, d\}$ are the grid indices, and associate each grid vertex with the image features at its projected image positions. We then apply 5 layers of 3D convolutions with filter size $3\times3\times3$ to the feature volume, and use an MLP to decode the TSDF value at each sampled grid vertex from its feature. In this way, we can compute the TSDF values at all the grid vertices $\{ \mathbb{X}_{whd} \}$ as,
\begin{equation}
	\mathbb{D} (\mathbb{X}_{whd}) = f (\mathbb{X}_{whd}, \mathbf{F}^{3D}( \mathbb{X}_{whd})) = s, \qquad s \in [-1,1].
	\label{equation:3D_discrete}
\end{equation}
Note that, Equation~\ref{equation:3D_discrete} is a discretized version of Equation~\ref{equation:3Dfeature} and only computes TSDF values for the pre-sampled volume grids. 
As we discuss in the supplementary file, we empirically find this discrete approach is close to the original continuous version with high-resolution feature volumes. The MLP $f(\cdot)$ can also be heavily simplified to just one layer in our experiments.

Alternatively, as shown in Figure~\ref{fig:toy_ablation} (a), we might use the 2D image features directly as input to the MLP to decode the TSDF values at grid vertices $\{ \mathbb{X}_{whd} \}$. To ensure a fair comparison, we employ additional $3\times3$ 2D convolutions in the image space to make the number of learnable parameters similar.







We train these two toy networks using pre-sampled volume grids at 256 resolution on the Twindom ~\cite{twindom} dataset, and evaluate them on the 160 testing human models. Table~\ref{tab:ablation_palvo} shows the Chamfer distance and point-2-surface errors of both methods. It becomes evident that learning a 3D feature volume with 3D convolutions leads to more accurate reconstructions\footnote{Note that PIFu~\cite{iccv2020PIFu} reconstructs a continuous surface which is not limited to the pre-sampled grid vertices $\{ \mathbb{X}_{whd} \}$. In the same experiment, its Chamfer distance error is 0.592 and P2S error is 0.538, which are slightly better than our discretized version with 2D features, but inferior to the version employing 3D features.} since the 3D CNNs can better leverage context information.



\begin{figure*}[]
\begin{center}
    \includegraphics[width=1.0\linewidth]{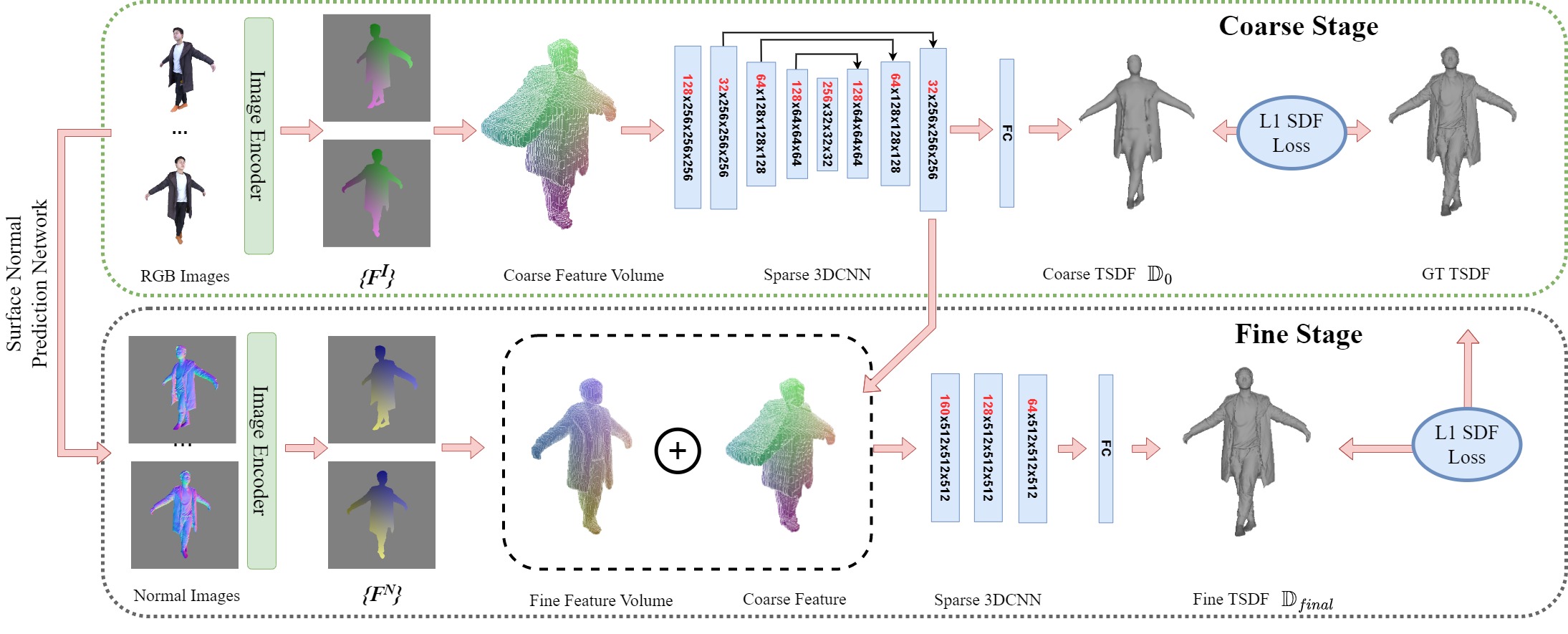}
\end{center}
\vspace{-0.15in}
    \caption{Pipeline of our shape reconstruction: given sparse multi-view images, our method works in a coarse to fine manner to compute the TSDF volume of the clothed human.
    During the coarse stage, we gather  features from the input images for voxels within the visual hull and compute a coarse TSDF $\mathbb{D}_0$ by a sparse 3D CNN. During the fine stage, we discretize a narrow band nearby the surface computed at the coarse level, and collect normal features to compute a fine TSDF $\mathbb{D}_{\text{final}}$ using another sparse 3D CNN.}
\label{fig:pipeline}
\vspace{-0.15in}
\end{figure*}

\begin{table}[]
\small
\begin{tabular}{|c|c|c|c|c|c|c|}
\hline
            & 64F5M  &   128F5M
            & 256F5M &   256F5M*  &
            256F1M* &
            256F'5M \\ \hline
Chamfer   & 0.574      &  0.459   & 0.406        & 
0.404   &   0.432 &   0.592             \\ \hline
P2S       & 0.524      &
0.395   & 0.332        &
0.358   & 0.365 & 0.538\\
\hline
\end{tabular}
\caption{
Results of the toy network using various network settings. In each column, the F-number represents  the volume resolution, while the M-number denotes  the depth of the MLP decoder. M* signifies  a discrete MLP which  evaluates TSDF values only on the grid vertices, and F' indicates  convolution is not applied to the 3D features.}
\label{tab:resolution_ablation}
\vspace{-0.2in}
\end{table}


\subsection{Network Settings}
Our formulation in Equation~\ref{equation:3D_discrete} includes a convolutional encoder and an MLP decoder, similar to the hybrid representation in \cite{chibane20ifnet,Peng2020ECCV}. In this subsection, we explore variations in the network settings, including feature volume resolution, discrete versus continuous MLP, and the depth of MLP, to understand their impact on shape reconstruction results.
We first test our toy network with 3D features using different volume resolutions. In these experiments, we choose to learn a continuous surface represented by the MLP $f(\cdot)$, as in PIFu~\cite{iccv2020PIFu}.
Specifically, we randomly sample 3D points and trilinearly interpolate features at these sampled points from the discrete grid vertices $\{ \mathbb{X}_{whd} \}$, and then use the MLP $f(\cdot)$ to compute the TSDF value. 
Table~\ref{tab:resolution_ablation} summarizes the results of various settings, where the F-number and M-number in each column represent  the volume resolution and the MLP depth, respectively. From the left three columns, it is evident  that increasing the volume resolution from $64$ to $256$ can significantly reduce reconstruction errors by about $30\%$, indicating that a high-resolution feature volume is crucial for precise results.

We further test other network settings. 
The two columns with an M* indicate results with a discrete MLP, which computes TSDF results only on the grid vertices $\{ \mathbb{X}_{whd} \}$. From these two columns, it is apparent that the discrete MLP only slightly compromises result quality, and even a 1-layer discrete MLP can achieve satisfactory  results. 
The rightmost column with an F' represents  results where convolution is not applied to the 3D feature volume. In this scenario, even a high-resolution 3D feature volume produces a substantial  error, which clearly highlights the effectiveness of 3D convolution.

\subsection{Quantization Error}\label{sec:quantization}
The formulation in Equation~\ref{equation:3D_discrete} involves discretization, which is often undesirable and motivates implicit function representation~\citep{Mescheder2019,Park2019,Chen2019}. In the following, we analyze the quantization error of the TSDF $\mathbb{D}$ with different volume resolutions. Surprisingly, we find that with relatively high volume resolution, e.g. 512 or higher, the quantization error is no longer a limiting factor for the reconstruction accuracy of clothed humans. 
Specifically, we compute ground truth TSDFs according to the ground truth mesh models in the THuman2.0~\cite{zheng2021deepmulticap} dataset. We then discretize those TSDFs into volumes of different resolutions, ranging from $32\times32\times 32$ to $1024\times1024\times1024$. The quantization error is measured by the average error in TSDF values on the ground truth mesh surface. As shown in Figure~\ref{fig:quantization_error}, the quantization error drops quickly with higher volume resolution. 
When the volume resolution is 512 or higher, the quantization error is less than 0.05 cm, much smaller than the reconstruction error of SOTA methods~\cite{iccv2020PIFu,zheng2021pamir,zheng2021deepmulticap}, which typically exceeds 0.5 cm. This indicates that  discrete TSDF representation is not a limiting factor for clothed human reconstruction with a volume resolution of 512 or higher.

\section{Method}
As discussed in Section~\ref{sec:motivation}, a convolutional encoder with 3D feature volume can boost shape reconstruction accuracy. While discretization causes additional quantization errors, a high-resolution volume can effectively mitigate this issue. Therefore, it is important to design an efficient method to overcome the memory and computation costs associated with high-resolution volumes for improved results. For this purpose, we design a sophisticated voxel culling process, implement a coarse-to-fine strategy, and employ the efficient submanifold sparse convolutional networks~\cite{3DSemanticSegmentationWithSubmanifoldSparseConvNet}.

Figure~\ref{fig:pipeline} shows our system pipeline for shape reconstruction. Our method works in two stages from coarse to fine. In the coarse stage, we adopt a $256\times 256\times 256$ resolution volume and employ the visual hull of the foreground object to eliminate irrelevant voxels. 
The remaining voxel grids are associated with image features at their projected positions. We then apply the efficient subspace sparse 3D CNN~\cite{3DSemanticSegmentationWithSubmanifoldSparseConvNet} to compute an initial TSDF, $\mathbb{D}_0$. Unlike conventional 3D CNN, sparse 3D CNN builds a hash-table for indexing non-zero elements and the convolution operator only applies to those non-zeros elements, which makes the convolution computational and memory efficient when the input tensor is sparse. In the fine stage, we focus on the narrow band nearby the zero-level set of $\mathbb{D}_0$ and discretize that narrow band into smaller voxels of $512\times 512\times 512$ resolution.
Each voxel grid is then associated with features from normal maps computed from the input images by the method~\cite{hourglass16}. We further fuse a coarse geometry feature from the coarse level, and apply the sparse 3D CNN again on the fine volume to compute the final TSDF, $\mathbb{D}_{\text{final}}$. The visual hull culling and narrow-band culling substantially  reduce the sampling grids, making the feature volume sparse enough for efficient sparse convolution.

After reconstructing the 3D shape, we proceed to estimate the surface texture. Texture maps need even higher resolution to capture appearance details. To address this problem, instead of na\"{\i}vely applying our TSDF regression network to compute a color volume that evaluates color as a function of coordinates, we choose to solve a field of blending weights. At each 3D point, the surface color is the weighted average of the colors at its image projections. We only solve the blending weights for a narrow band nearby the final surface $\mathbb{D}_{\text{final}}$ for better efficiency. Our pipeline to solve this blending weight volume is shown in Figure~\ref{fig:pipe_color}.

\subsection{Feature Volume Construction}\label{sssec:featurevolume}
To construct the feature volume, we initial a cubic volume grid $\mathbb{V}$ with an edge length of $256$cm and a resolution  of ${256}\times{256}\times{256}$. We then project each voxel grid point $\mathbb{V}_{whd}$ on the input mask images $\{{M}_i\}$ to discard points outside of the visual hull. Pruning by the visual hull significantly reduce the number of `active' voxel vertices in the volume. Typically, over $95\%$ of the voxel grids are culled away by the visual hull, leaving around 200--300K voxels remaining. Given the set of remaining voxel grids $\{ \mathbb{V}_{whd} \}$, we project them onto feature maps to compute a feature on each grid vertex as follows,
\begin{equation}
	\mathbb{F}^I_{whd} = Mean({F}^{I}_i(\Pi_i (\mathbb{V}_{whd}))).
	\label{equation:projection}
\end{equation}
Here, $\Pi_i$ is the perspective projection of the input image ${I}_i$. We sample the 2D feature maps ${F}^I_i$ using bi-linear interpolation at the projected positions, and average the sampled features from all views to compute the feature volume $\mathbb{F}^I$. The image feature ${F}^I_i$ is computed from the input image ${I}_i$ by a single stacked hourglass network~\cite{hourglass16}, which has 128 feature channels and at resolution of $256\times256$. 

\subsection{Coarse to Fine Reconstruction}

To ensure the memory consumption and inference speed, we take a coarse-to-fine architecture to compute the TSDF $\mathbb{D}$. 

\paragraph{Coarse stage}
We use a 3D U-Net with skip layers to encode the topology context. As shown in Figure~\ref{fig:pipeline} the 3D U-Net consists of three conv and deconv blocks with skip connections. The initial TSDF $\mathbb{D}_0$ at each volume grid is computed by an FC layer, i.e. the MLP $f(\cdot)$ in Equation~\ref{equation:3D_discrete}. More network details are provided in the supplementary file. We have experimented with more FC layers and empirically found that adding more layers does not help, thanks to the 3D convolutional feature encoder with proper local information encoding.
This network outputs a coarse TSDF $\mathbb{D}_0$ with the same resolution as the feature volume $\mathbb{F}^I$. 
During training, we calculate the ground truth TSDF for each clothed human model from the ground truth mesh model. We further truncate the TSDF value within [-5cm,5cm]. The training loss function for the coarse stage is then defined as:
\begin{equation}
	L_c = \sum_{(w,h,d) \in \mathcal{V}_{\text{hull}} }
	 \left \| (\mathbb{D}^{\text{pred}}_{whd} + bias_c) - \mathbb{D}^{\text{gt}}_{whd} \right \|_{L1} 
	\label{equation:coarse_loss}
\end{equation}
Here, $\mathbb{D}^{\text{pred}}$ and $\mathbb{D}^{\text{gt}}$ are the predicted and ground truth TSDF volume respectively, and $\mathcal{V}_{\text{hull}}$ is the set of remaining voxel grids after visual hull culling. 
The training loss is the L1 distance between the predicted and ground truth TSDF values. 
Note that the predicted TSDF values of voxel grids outside the visual hull will be zero according to the submanifold sparse convolution (SSC). Hence, we add a constant bias $bias_c = 0.05$m which is the truncation distance to the TSDF.

\begin{figure}
\begin{center}
    \includegraphics[width=1.0\linewidth]{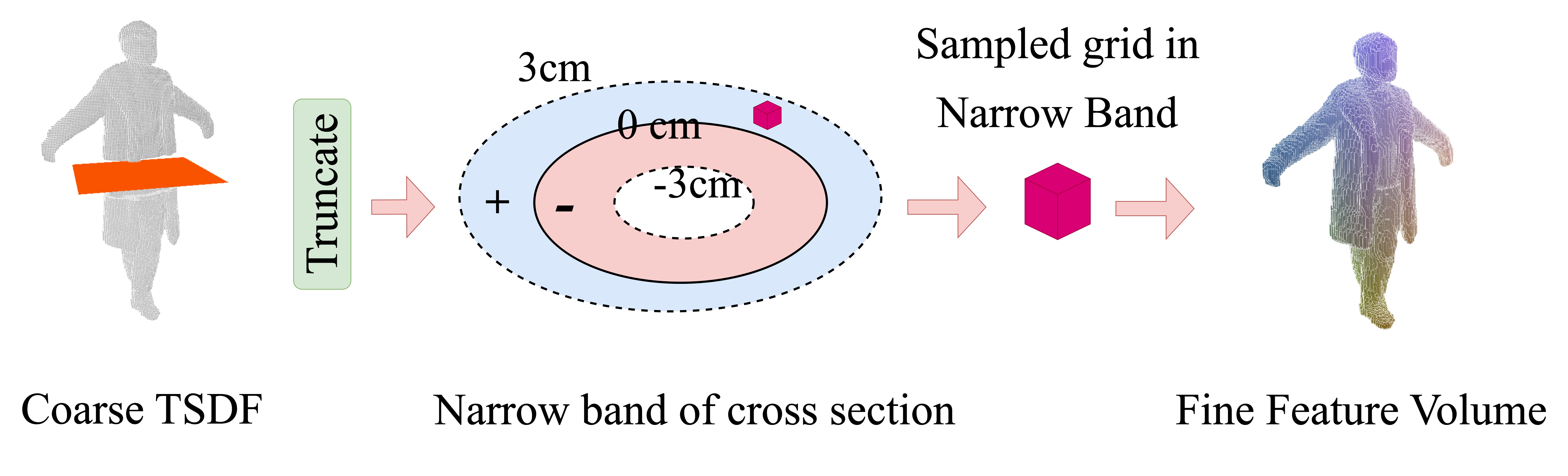}
\end{center}
\vspace{-0.15in}
    \caption{At the fine stage, we focus on a narrow band nearby the coarse stage result and discretize it to voxels. We then project each voxel vertex to the normal feature map to form the fine feature volume $\mathbb{F}^f$.}
\label{fig:c2f}
\vspace{-0.15in}
\end{figure}

\paragraph{Fine stage}
After getting the coarse TSDF $\mathbb{D}_0$, we use another branch to further refine geometry details. At this stage, we down-sample the volume to denser voxels and associate each voxel vertex with high frequency shape information encoded in normal maps like~\cite{zheng2021deepmulticap,saito2020pifuhd}. To facilitate computation, we focus on a narrow band nearby the zero-level set of $\mathbb{D}_0$.
Specifically, we  tri-linearly interpolate the coarse TSDF volume $\mathbb{D}_0$ by $2$ times to a $512\times512\times512$ voxel grid. We only preserve all the voxel vertices satisfying: $ \left | \mathbb{D}_0 \right | < 0.03m$, which are within a narrow band of 6cm width around the zero-level set surface of $\mathbb{D}_0$. Figure~\ref{fig:c2f} shows the narrow band for re-sampling.
This narrow band removes irrelevant voxel vertices based on the initial result, helping to further improve the storage and computation efficiency of our method.

We compute a feature at each voxel vertex in the fine stage  from the normal feature maps as follows,
\begin{equation}
	\mathbb{F}^N_{whd} = Mean({F}^N_i(\Pi_i (\mathbb{V}_{whd}))).
	\label{equation:fine_projection}
\end{equation}
We use the same normal estimator proposed in ~\cite{zheng2021deepmulticap} to estimate the normal image $N_i$ for each input view $I_i$, and the normal feature map ${F}^N_i$ is computed from the input normal image ${N}_i$ by the same hourglass network as ${F}^I_i$. Furthermore, this feature is concatenated with the down-sampled coarse level features $\mathbb{F}^I_{whd}$ at the last two convolution layers, which have large receptive fields and encode strong shape information.
Since the fine stage mainly focuses on local shape details, we use a shallow sparse 3D CNN which has 3 Conv blocks followed by an FC layer to regress the final TSDF volume. More details of network architecture are in the supplementary file.

We define the training loss for the fine stage as follows,
\begin{equation}
	L_f = \sum_{(w,h,d) \in \mathcal{V}_{\text{band}}}
	\left \| ( \mathbb{D}^{\text{pred}}_{whd} + bias_f) - \mathbb{D}^{\text{gt}}_{whd} \right \|_{L1}.
	\label{equation:fine_loss}
\end{equation}
Here, $\mathcal{V}_{\text{band}}$ is the set of voxel grids within the narrow band. We also add a constant bias $bias_f = 0.03m$ to deal with vertices outside of the narrow band. 

\subsection{Texture Prediction}
With the shape reconstructed, we then estimate the color at each surface point. Instead of solving a color field encoded by an implicit function like the earlier works\cite{iccv2020PIFu,shao2022doublefield}, we exploit high-resolution input images for rich appearance details. 
Specifically, we estimate a blending weight vector $\mathbf{W}$ at each surface point $\mathbf{X}$. The color at $\mathbf{X}$ is then computed as a weighted average of the colors at its image projections,
\begin{equation}
    c(\mathbf{X}) = \sum_i \mathbf{W}_i \mathbf{I}_i (\mathbf{x_i}),
    \label{equ:color}
    \vspace{-0.1in}
\end{equation}
where $\mathbf{x}_i = \Pi_i(\mathbf{X})$ is the projection of $\mathbf{X}$ in image $\mathbf{I}_i$. Similar to shape reconstruction, we also sample a volume grid $\mathbb{W}$ and estimate the blending weights at the grid vertices $\mathbb{W}_{whd}$. In this way, our method essentially estimates a blended texture map over the surface, instead of computing a color field which tends to be limited by the 3D sampling rate. With our method, the predicted texture map carries sharp appearance details inherited from the input images.





\begin{figure*}
\begin{center}
    \includegraphics[width=1.0\linewidth]{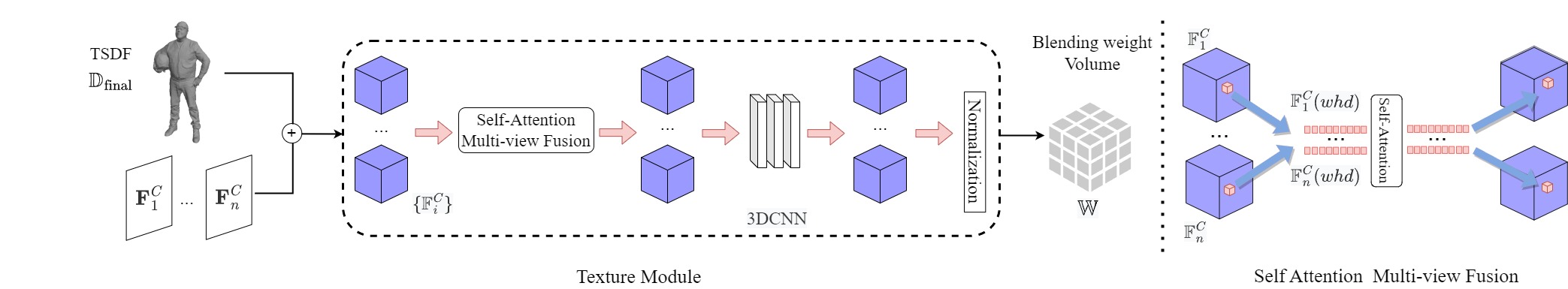}
\end{center}
\vspace{-0.15in}
    \caption{The texture module predicts a blending weight volume using the input texture feature maps $\mathbf{F}^C_i$ and the estimated TSDF $\mathbb{D}_{\text{final}}$.
    The final texture map is computed by interpolating pixel values from the input images according to the blending weights.}
\label{fig:pipe_color}
\vspace{-0.15in}
\end{figure*}



The network architecture of our texture weight estimation is shown in Figure~\ref{fig:pipe_color}. Thanks to the precisely reconstructed TSDF $\mathbb{D}_{\text{final}}$ from the shape branch, we only consider a 2cm width narrow-band nearby its zero-level set, which is defined as $ \left | \mathbb{D}_{\text{final}} \right | < 0.01m$. For each input image, we compute its texture feature ${\mathbf{F}^C_i}$ and construct a volume ${\mathbb{F}^C_i}$ by projecting them back to the discretized narrow-band. 
The texture feature maps $\mathbf{F}^C_i$ are also computed by an hourglass ~\cite{hourglass16} network from the input image $\mathbf{I}_i$ with a smaller network to extract a 32-channel feature of size $256\times256$. We further compute the truncated PSDF~\cite{tao2021function4d}, which is a view-dependent function indicating if the surface is viewed from a slanted direction, and concatenate it to the texture feature. 


The attention model~\cite{attention} is used here to handle visibility by re-weighting features across different views. Ideally, if a 3D point is not visible from a particular view, the projected color from that view should have less influence on the final blended color. Therefore, the attention module is used to adjust the contribution of the projected texture features by re-weighting them. Following this, we apply a sparse 3D CNN on these re-weighted feature volumes individually to regress the blending weight volume of each view. We then normalize these blending weight volumes across views by a soft-max to obtain the normalized blending weights $\mathbb{W}$.


The training loss for the texture blending weight estimation is defined as:
\begin{equation}
	L_c = \sum_{(w,h,d) \in \mathcal{W}_{\text{band}}}
	\left \|  \mathbb{C}^{\text{pred}}_{whd}  - \mathbb{C}^{\text{gt}}_{whd} \right \|_{L1}.
	\label{equation:color_loss}
     \vspace{-0.1in}
\end{equation}
Here, $\mathbb{C}^{\text{pred}}_{whd}$ is the surface color volume computed by applying the blending weights  $\mathbb{W}_{whd}$. The ground truth color volume $\mathbb{C}^{\text{gt}}_{whd}$ is generated from the ground truth textured mesh with nearest search. The set $\mathcal{W}_{\text{band}}$ includes all voxel grids within the narrow band for color estimation.

\begin{figure}
\begin{center}
    \includegraphics[width=1.0\linewidth]{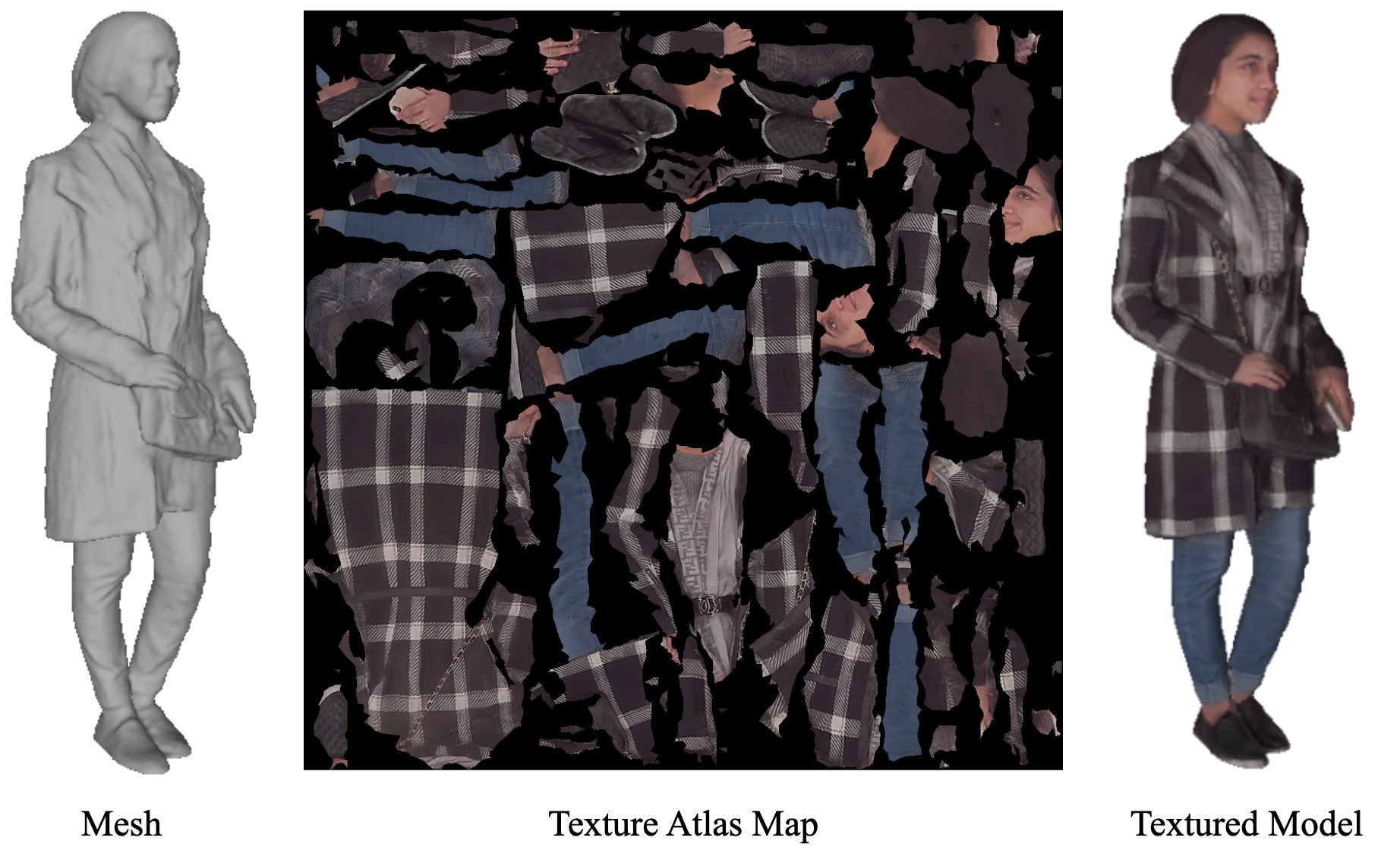}
\end{center}
\vspace{-0.15in}
    \caption{The estimated shape, texture atlas map, and a rendering of the textured model. }
\label{fig:atlas}
\vspace{-0.15in}
\end{figure}

After solving the blending volume $\mathbb{W}$, we use the Blender ~\cite{Blender} to generate a texture atlas map of the reconstructed mesh model. For each pixel in the atlas map, we use barycentric interpolation to compute its 3D location from the mesh vertices, and then determine its color according to Equation~\ref{equ:color}. Figure ~\ref{fig:atlas} shows the atlas map computed by our method. To capture rich details in the input images, a high-resolution texture atlas map, such as 2K resolution, can be chosen. In this way, our method generates high-quality textured results.

\begin{table*}[]
\begin{tabular}{ccccccccc}
\hline
\multicolumn{1}{l}{} & \multicolumn{2}{c}{1 view}             & \multicolumn{2}{c}{2 views}                 & \multicolumn{2}{c}{4 views}                 & \multicolumn{2}{c}{6 views}                 \\
\multicolumn{1}{l}{} & Chamfer              & P2S                  & Chamfer              & P2S                  & Chamfer              & P2S                  & Chamfer              & P2S                  \\ \hline
PIFu                 & 2.528/1.612          & 2.421/1.587          & 1.626/1.200          & 1.507/1.170          & 0.929/0.823          & 0.783/0.773          & 0.776/0.678          & 0.725/0.625          \\ \hline
PIFuHD               & 2.814/1.725          & 2.793/1.704          & 1.369/1.162          & 1.223/1.126          & 0.821/0.765          & 0.719/0.645          & 0.720/0.698          & 0.705/0.501          \\ \hline
DeepMultiCap         &     --               &                --               & 1.529/1.159          & 1.496/1.117          & 1.150/0.969          & 1.115/1.001          & 1.062/0.890          & 1.024/0.944          \\ \hline
Doublefield          &     --               &              --                 &     --               &              --                 & 0.836/0.905          & 0.822/0.869          & 0.711/0.779          & 0.690/0.740          \\ \hline
Ours(256)            & 2.457/1.563          & 2.374/1.537          & 1.221/0.899          & 1.080/0.860          & 0.810/0.550          & 0.629/0.500          & 0.668/0.459          & 0.470/0.402          \\ \hline
Ours(512)            & \textbf{2.398/1.565} & \textbf{2.363/1.539} & \textbf{1.110/0.889} & \textbf{1.052/0.837} & \textbf{0.514/0.447} & \textbf{0.429/0.389} & \textbf{0.390/0.314} & \textbf{0.287/0.242} \\ \hline
\end{tabular}
\caption{Mean Chamfer and point-2-surface (P2S) errors of the reconstructed mesh on the Twindom dataset. In each entry, we report two error metrics as $x/y$, where $x$ represents recall and $y$ stands for precision.}
\label{tab:comparison}
\vspace{-0.1in}
\end{table*}

\begin{table*}[]
\begin{tabular}{ccccccccc}
\hline
\multicolumn{1}{l}{} & \multicolumn{2}{c}{1 view}             & \multicolumn{2}{c}{2 views}                 & \multicolumn{2}{c}{4 views}                 & \multicolumn{2}{c}{6 views}                 \\
\multicolumn{1}{l}{} & Chamfer              & P2S                  & Chamfer              & P2S                  & Chamfer              & P2S                  & Chamfer              & P2S                  \\ \hline
PIFu                 & 1.975/1.540          & 1.872/1.511          & 1.370/1.216          & 1.249/1.169          & 0.893/0.653          & 0.742/0.604          & 0.660/0.523          & 0.472/0.462          \\ \hline
PIFuHD               & 2.142/1.936          & 2.121/1.916          & 1.140/0.915          & 0.998/0.873          & 0.739/0.589          & 0.564/0.523          & 0.630/0.492          & 0.438/0.433          \\ \hline
DeepMultiCap         &          --          &              --                 & 1.114/0.928          & 1.077/0.914          & 0.932/0.781          & 0.891/0.777          & 0.681/0.678          & 0.678/0.676          \\ \hline
Doublefield          &          --          &              --                 &          --          &              --                 & 0.743/0.788          & 0.621/0.748          & 0.652/0.664          & 0.579/0.621          \\ \hline
Ours(256)            & 1.922/1.516          & 1.824/1.485          & 0.995/0.735          & 0.847/0.691          & 0.644/0.412          & 0.445/0.345          & 0.570/0.341          & 0.356/0.275          \\ \hline
Ours(512)            & \textbf{1.852/1.515} & \textbf{1.819/1.480} & \textbf{0.822/0.689} & \textbf{0.763/0.602} & \textbf{0.453/0.360} & \textbf{0.372/0.296} & \textbf{0.348/0.271} & \textbf{0.252/0.195} \\ \hline
\end{tabular}
\caption{Mean Chamfer and point-2-surface (P2S) errors of the reconstructed mesh on the Multihuman dataset. In each entry, we report two error metrics as $x/y$, where $x$ represents  recall and $y$ stands for precision.}
\label{tab:comparison_multihuman}
\end{table*}

\section{Experiments}

\subsection{Implementation Details}
We experimented with three commonly used datasets, Twindom ~\cite{twindom}, THuman2.0~\cite{zheng2021deepmulticap}, and MultiHuman\cite{zheng2021deepmulticap}. These datasets consist of high-quality scanned 3D models of clothed humans with varying poses and body shapes. We followed \cite{iccv2020PIFu} to generate multi-view images under spherical harmonic lighting to train our network. Before training our shape networks, we pre-computed the ground truth TSDF $\mathbb{D}^{\text{gt}}$ for each human model in our training set. For  training the texture network, we computed the ground truth color volume $\mathbb{C}^{\text{gt}}$ by finding the nearest mesh vertex for each volume grid. In the experiment, we used 1,000 models from Twindom~\cite{twindom} and THuman2.0~\cite{zheng2021deepmulticap} for training. Another 200 models from Twindom and 30 models from MultiHuman were used for testing. We employed the Adam optimizer with a learning rate of 1e-4. The network was trained in an end-to-end fashion for 30 epochs, and the training of our pipeline took approximately 8 hours using 8 NVIDIA A100 GPUs. 

At testing time, for each model, we used 1/2/4/6 input images from different viewpoints to reconstruct the clothed human model. To estimate normal maps, we employed the pre-trained model from \cite{zheng2021deepmulticap}. Our testing experiments were performed with an NVIDIA 3090 GPU. The breakdown of  the running time for our method with 6 input images and 512 volume resolution is provided in Table~\ref{tab:shape_time_cost} and Table~\ref{tab:texture_time_cost} for shape and texture estimation respectively. The most time-consuming step of our shape reconstruction is to extract feature maps $\{ \mathbf{F}^I_i \}$ and $\{ \mathbf{F}^N_i \}$.  We can use buffer swapping techniques with two GPUs to achieve $\times$2 speedup. 

In terms of memory consumption, our method takes 18G and 45G of GPU memory during training for the 256 and 512 volume resolutions respectively. The testing time memory consumption is 12G and 18G.

\begin{table}[]
\small
\begin{tabular}{c|ccc|ccc|c}
\hline
     & \multicolumn{3}{c|}{Coarse Stage}                                       & \multicolumn{3}{c|}{Fine Stage}                                        & Total \\ \hline
     & \multicolumn{1}{c|}{$\{ \mathbf{F}^I_i$ \}} & \multicolumn{1}{c|}{VH}   & 3D CNN & \multicolumn{1}{c|}{$\{ \mathbf{F}^N_i$ \}} & \multicolumn{1}{c|}{NB}   & 3D CNN &       \\ \hline
Time (ms) & \multicolumn{1}{c|}{112}   & \multicolumn{1}{c|}{85} & 52   & \multicolumn{1}{c|}{112}   & \multicolumn{1}{c|}{11} & 93  & 465 \\ \hline
\end{tabular}
\caption{Shape reconstruction time cost. Here, columns $\{ \mathbf{F}^I_i \}$ and $\{ \mathbf{F}^N_i \}$ are the time on computing these feature maps. `VH' and `NB' are the time on visual hull culling and narrow band culling. }
\label{tab:shape_time_cost}
\vspace{-0.15in}
\end{table}


\begin{table}[]
\begin{tabular}{c|c|c|c|c|c}
\hline
     & $\{ \mathbf{F}^C_i \}$  & Attention & 3D CNN & UV Atlas & Total \\ \hline
Time (ms) & 28    & 18      & 150 & 57  & 253 \\ \hline
\end{tabular}
\caption{Texture reconstruction time cost. Here, the column $\{ \mathbf{F}^C_i \}$ indicates the time for computing the feature maps.} 
\label{tab:texture_time_cost}
\vspace{-0.15in}
\end{table}

\subsection{Quantitative Results on Synthetic Data}
In this subsection, we compare our methods against other reconstruction methods, including Multi-view PIFu~\cite{iccv2020PIFu}, Multi-view PIFuHD~\cite{saito2020pifuhd}, DeepMultiCap \cite{zheng2021deepmulticap}, DoubleField ~\cite{shao2022doublefield}, and DiffuStereo~\cite{shao2022diffustereo}.
We test the robustness  of these methods with different numbers of input images. To ensure a fair comparison, we implemented MultiView PIFuHD~\cite{saito2020pifuhd} and MultiView PIFu~\cite{iccv2020PIFu} based on the public code of their single-view versions. We used the same training and testing data for MultiView PIFu, MultiView PIFuHD, and our method. The authors of DeepMultiCap~\cite{zheng2021deepmulticap} and DoubleField~\cite{shao2022doublefield} kindly provided us with their results. According to their paper, these two methods were trained on a larger set of data than our method. To make the comparison fair, we sample a 512$\times$512$\times$512 volume to compute the final shape using the marching cube algorithm in all the compared methods.

In the case of a single input image, we determine the visual hull by truncating a cone defined by the camera center and the image silhouette with depth thresholds of $-0.5$m and $0.5$m.
We retrain the public code of PIFu~\cite{iccv2020PIFu} on our dataset. PIFuHD~\cite{saito2020pifuhd} only releases testing code, so we test its pre-trained model on our dataset.
 
\begin{table}[]
\begin{tabular}{l|l|l}
\hline
            & DiffStereo  & Ours        \\ \hline
Chamfer/P2S & 0.120/0.126 & 0.158/0.103 \\ \hline
\end{tabular}
\caption{The shape precision of our method and the DiffuStereo\cite{shao2022diffustereo} on the 8-view setting.}
\label{tab:compare_diffstereo}
\vspace{-0.15in}
\end{table}

Table~\ref{tab:comparison} and Table~\ref{tab:comparison_multihuman} provide a quantitative comparison of different methods with 1--6 input images. Note that Deepmulticap~\cite{zheng2021deepmulticap} cannot work with a single input image, while DoubleField~\cite{shao2022doublefield} cannot work with a single or two input images. Thus, their results are absent for those settings. 
We report our results with two different volume resolutions, 256 and 512 for the fine stage (with corresponding coarse stage resolutions of 128 and 256, respectively). To better evaluate these methods, we report both recall and precision errors for all  methods. Precision is computed as the average distance between each vertex on the predicted mesh and its nearest correspondence on the ground truth mesh. Conversely, recall is the average distance between each vertex on the ground truth mesh and its nearest correspondence on the predicted mesh. 
Our method, with a 512 volume resolution, has the lowest errors among all methods for different numbers of input images. On the Twindown dataset (shown in Table~\ref{tab:comparison}), our method reduces the mean point-2-surface (P2S) precision to $0.24$cm  with $6$ input images, a remarkable error reduction of $51\%$ over DoubleField~\cite{shao2022doublefield} and Multi-view PIFu~\cite{iccv2020PIFu,saito2020pifuhd}. Furthermore, it achieves a mean P2S precision of $0.39$cm using just $4$ input images, with over a $39\%$ error reduction. Even with only $2$ input images, our method can achieve a $0.84$cm mean P2S precision, surpassing  DeepMultiCap~\cite{zheng2021deepmulticap} with 6 input views. This significant  improvement over SOTA methods is also apparent on the MultiHuman dataset (shown in Table~\ref{tab:comparison_multihuman}) and with the Chamffer error metric. 
When using a single input image, the performance improvement by our method is smaller. This is reasonable since the main source of error here is the single view depth ambiguity, which cannot be solved by our 3D convolution without additional input images.

\begin{figure}
\begin{center}
    \includegraphics[width=1.0\linewidth]{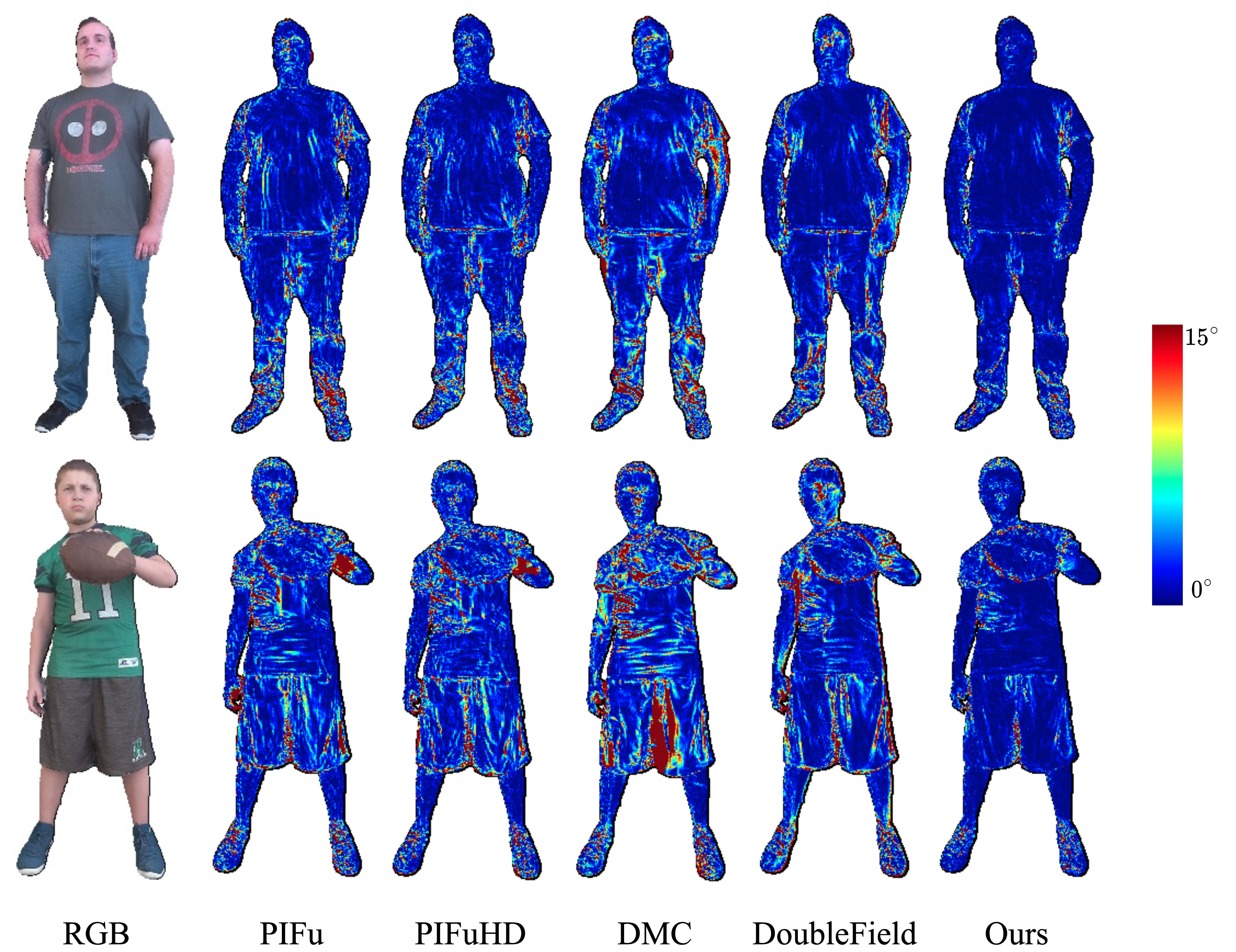}
\end{center}
\vspace{-0.15in}
    \caption{The normal errors of different methods are visualized as heat maps. }
\label{fig:normal_compare}
\vspace{-0.1in}
\end{figure}

DiffuStereo~\cite{shao2022diffustereo} requires image pairs with a smaller baseline for their diffusion-based stereo matching, which does not work on our sparse view setting in Table~\ref{tab:comparison} and Table~\ref{tab:comparison_multihuman}. Table~\ref{tab:compare_diffstereo} compares our method with DiffuStereo using the same 8-view setting as \cite{shao2022diffustereo}, where each view has an adjacent view facilitating stereo reconstruction. We employ our model trained for the 6-view input, which has not been trained or finetuned on the 8-view setting. Our P2S precision is 18\% smaller than that of DiffuStereo, which demonstrates the generalization capability of our method.  In this case, we follow \cite{shao2022diffustereo} to normalize the height of all human subjects to 1 meter when evaluating the error metrics, resulting in smaller error metrics than those in Table~\ref{tab:comparison} and Table~\ref{tab:comparison_multihuman}.

\begin{table}[]
\begin{tabular}{l|l|l|l|l|l}
\hline
           & PIFu   & PIFuHD & DMC & DoubleField & Ours  \\ \hline
Twindom    & 9.76  & 9.69  & 13.02       & 11.18      & \textbf{6.94} \\ \hline
Multihuman & 10.61 & 10.46 & 11.92       & 11.66      & \textbf{7.25} \\ \hline
\end{tabular}
\caption{The normal errors of different methods measured by the mean angular error (in degrees). Here DMC stands for the `DeepMultiCap' method. }
\label{tab:normal_error}
\vspace{-0.15in}
\end{table}

Table~\ref{tab:normal_error} and Figure~\ref{fig:normal_compare} show the surface normal error of different methods with 6 input views to evaluate their capability in capturing fine-scale shape details. To compute the surface normal error, we re-render the reconstructed surface into normal maps and compare them with the ground truth results. Our method achieves the smallest mean angular error on both datasets, demonstrating our capability of reconstructing shape details. As shown in Figure~\ref{fig:normal_compare}, other methods often produce larger errors at concave regions, such as the pants in the second row.

Table~\ref{tab:color_comparison} assesses the rendering quality of our textured models using  PSNR and SSIM metrics. All results are obtained under the 6-view setting. We cite the results of PixelNerf~\cite{yu2020pixelnerf} and DoubleField~\cite{shao2022doublefield} from DoubleField~\cite{shao2022doublefield}. 
Our method achieves the highest score on both metrics. 

\begin{figure*}
\begin{center}
    \includegraphics[width=1.0\linewidth]{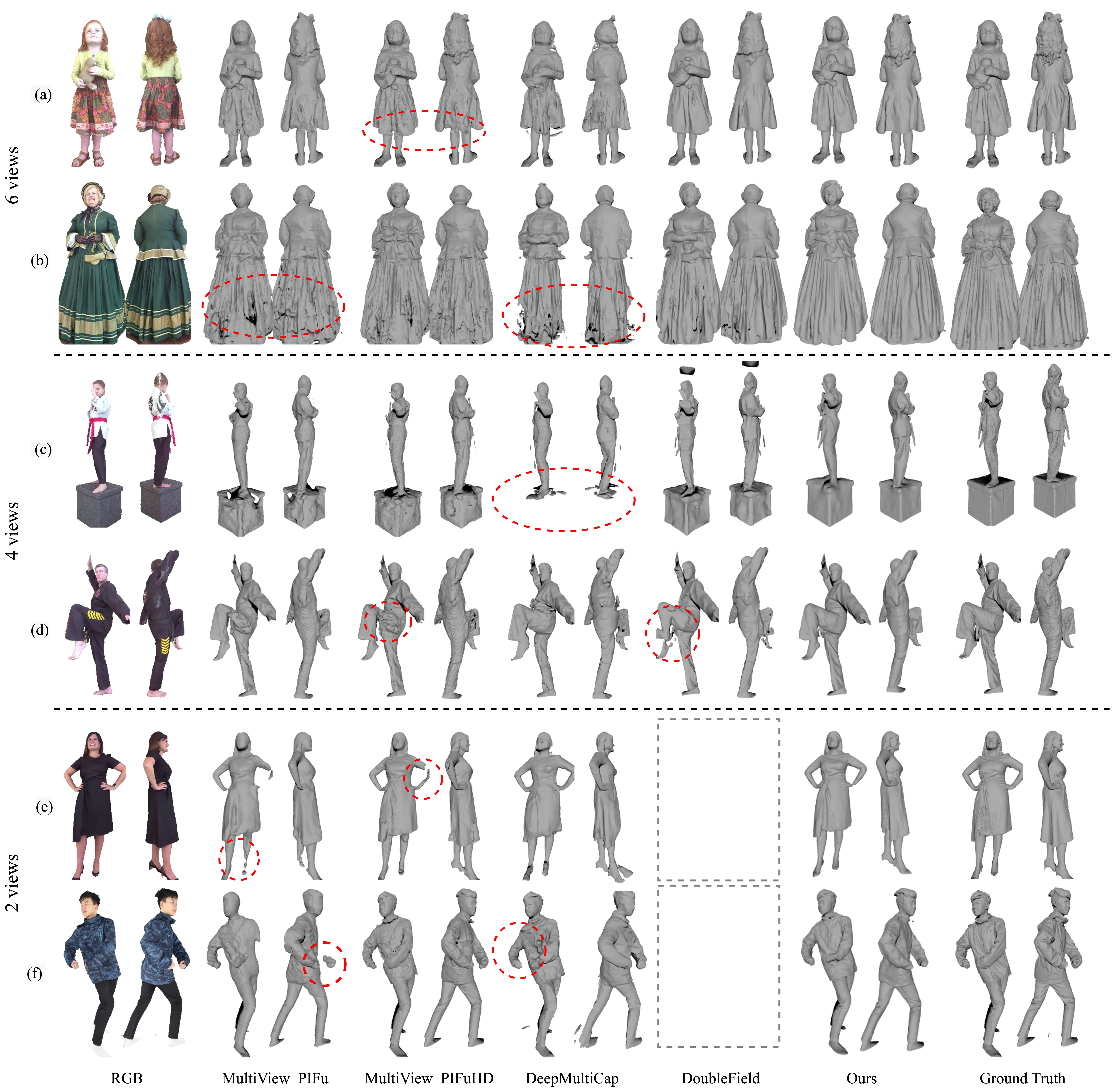}
\end{center}
\vspace{-0.15in}
    \caption{Visual results on the synthetic dataset. From left to right, shown are one of the input images, results from Multiview-PIFu~\cite{iccv2020PIFu}, Multiview-PIFuHD~\cite{saito2020pifuhd},
    DeepMulticap~\cite{zheng2021deepmulticap}, DoubleField~\cite{shao2022doublefield}, our method, and ground truth, respectively. }
\label{fig:compare}
\vspace{-0.15in}
\end{figure*}


\begin{table}[]
\begin{tabular}{l|l|l|l|l}
\hline
     & PIFu  & PixelNerf & DoubleField & Ours  \\ \hline
PSNR & 20.66 & 21.85     & 23.56   & \textbf{26.31} \\ \hline
SSIM & 0.807 & 0.813     & 0.857   & \textbf{0.863} \\ \hline
\end{tabular}
\caption{PSNR and SSIM of the re-rendered mesh on the synthetic dataset. Our method produces results most consistent with the ground truth.}
\label{tab:color_comparison}
\vspace{-0.15in}
\end{table}



\begin{table}[]
\begin{tabular}{c|cc|cc}
\hline
                   & \multicolumn{2}{c|}{\begin{tabular}[c]{@{}c@{}}Twindom \end{tabular}} & \multicolumn{2}{c}{\begin{tabular}[c]{@{}c@{}}MultiHuman\end{tabular}} \\
                   & Chamfer                                       & P2S                                        & Chamfer                                   & P2S                                   \\ \hline
AB1 (G)              &     0.351                                          &                 0.286                           &    0.303                                      &              0.233                         \\ \hline
AB2 (N)                  &     0.339                                          &                 0.275                           &                        0.280                   &          0.207                             \\ \hline

Proposed method                &         0.314                                      &             0.242                               &     0.271                                      &       0.195                                \\ \hline
\end{tabular}
\caption{Results of different ablation settings in shape reconstruction. AB1 and AB2 use different input features at the fine stage. }
\label{tab:Ablation_shape}
\vspace{-0.15in}
\end{table}



\begin{figure*}
\begin{center}
    \includegraphics[width=0.95\linewidth]{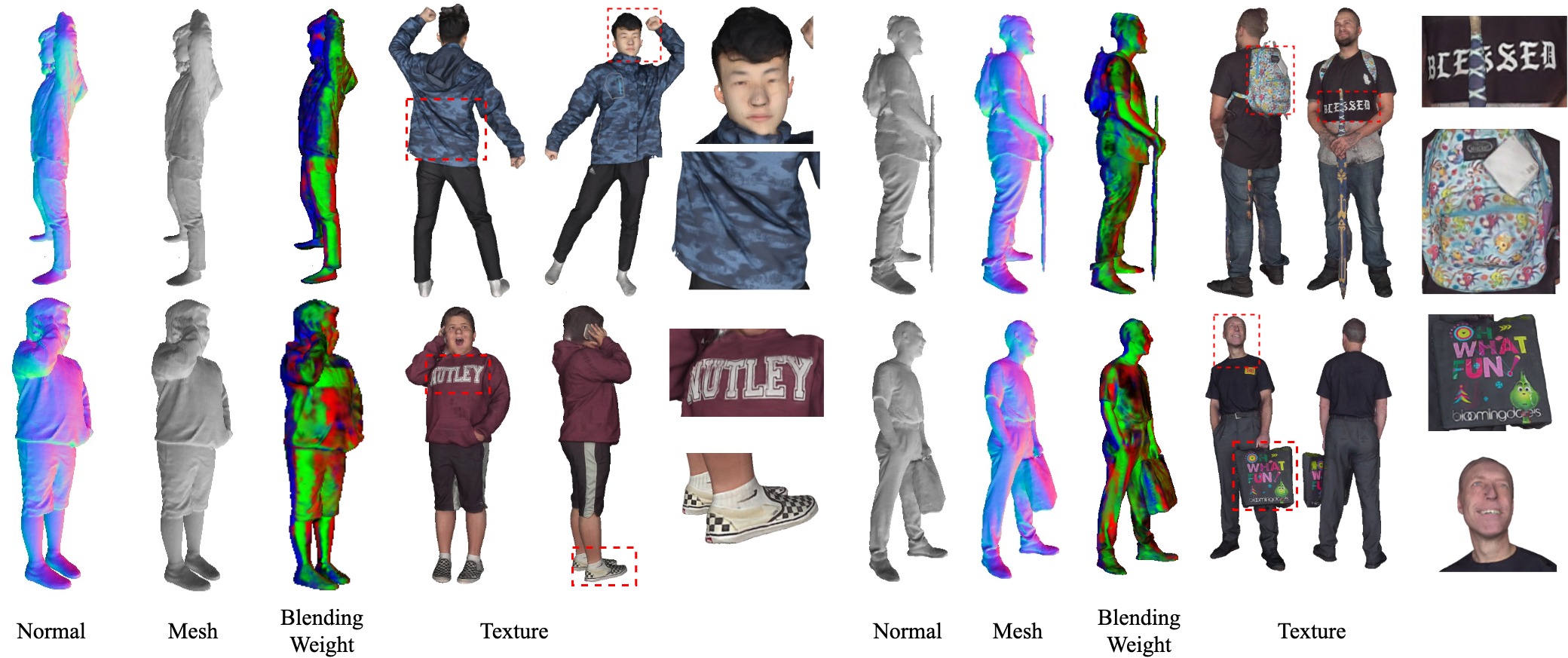}
\end{center}
\vspace{-0.15in}
    \caption{Texturing results. The blending weight map shows the weights of three input images in the RGB channels. The final texture map captures appearance details, such as facial expressions and cloth patterns, as shown in the zoomed-in regions. }
\label{fig:texture}
\vspace{-0.15in}
\end{figure*}

\begin{figure*}
\begin{center}
    \includegraphics[width=0.95\linewidth]{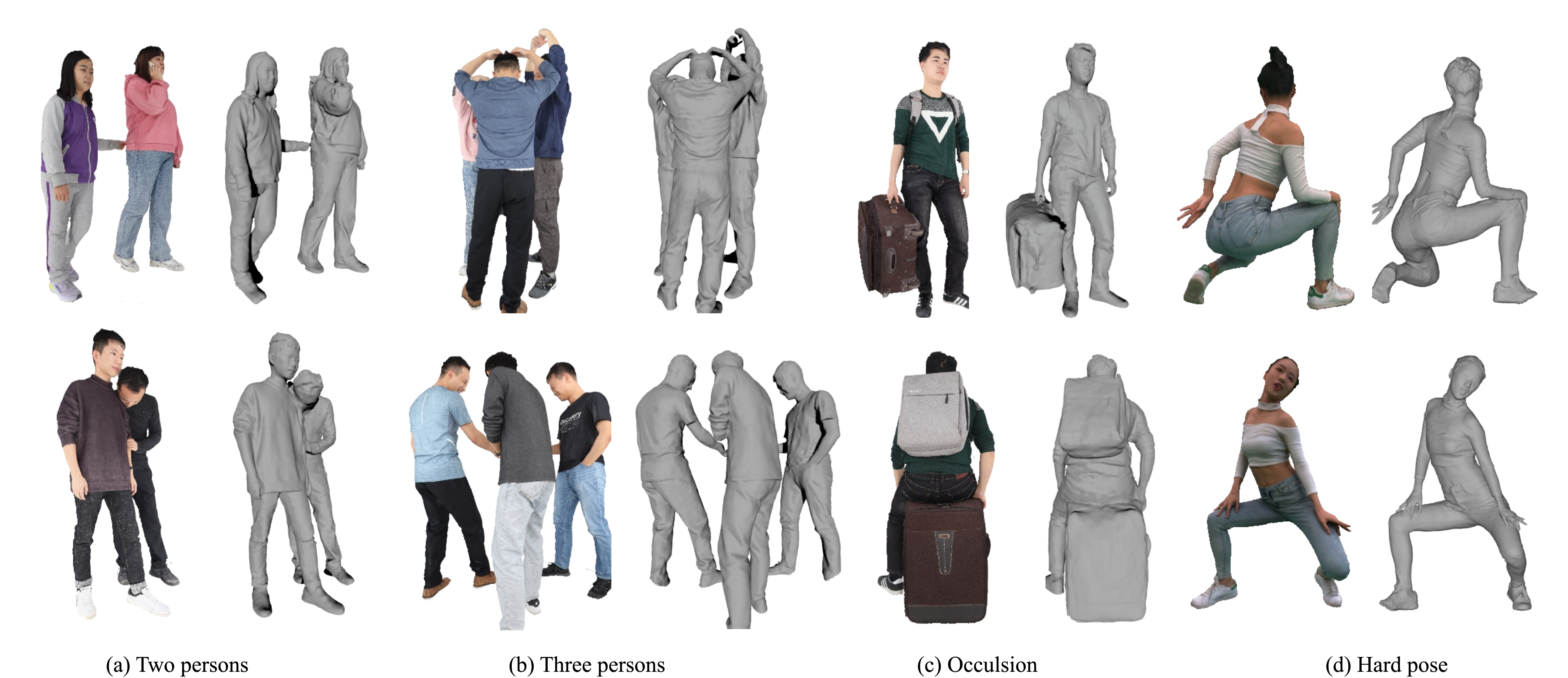}
\end{center}
\vspace{-0.15in}
    \caption{Some challenging cases on synthetic data. Our method is robust to multiple persons (without instance segmentation), occlusion, and rare poses.}
\label{fig:challenging}
\vspace{-0.15in}
\end{figure*}

To better understand the quantitative comparison, we visualize some of the results in Figure~\ref{fig:compare}, where (a)--(b), (c)--(d), and (e)--(f) are results reconstructed with 6, 4, and 2 input images, respectively. From left to right, the shown figures are input images, results from Multi-view PIFu, Multi-view PIFuHD, DeepMultiCap, DoubleField, our method, and ground truth, respectively. It is evident that our method generates more shape details and is more robust to loose garments and rare poses. From examples (e, f), we can observe that Multi-View PIFu and Multi-View PIFuHD often generate broken arms when the number of input images is small, while our method does not suffer from this problem. Examples (a, b) highlight our strength in handling  loose garments, where other methods often generate noisy reconstructions. Examples (c, d) showcase our capability in addressing rare poses and unusual  objects.


Figure~\ref{fig:texture} visualizes the recovered normal maps and blending weight maps for some examples. We visualize the blending weights of three input images in the respective RGB channels. The smooth transition of these weights generates seamless textured models with vivid texture details, as shown in the zoomed-in regions.

Figure~\ref{fig:challenging} shows some challenging examples, such as occlusion and multiple persons. Our method can still recover faithful shape details and poses in these situations. Note that we use the ground truth foreground segmentation, which includes the luggage and all persons together. Instance segmentation, as employed in DeepMultiCap ~\cite{zheng2021deepmulticap}, is not used here. Surprisingly, our method can even reconstruct backpacks and luggage quite  well, even though it has never been trained on these types of objects. We believe our 3D feature volume helps to learn local implicit functions, like those in \cite{Jiang2020}, which generalize well across object categories. This is because, locally, backpacks and luggage have similar shapes as clothed humans. More animated examples are provided in the supplementary video.

\begin{figure*}
\begin{center}
    \includegraphics[width=1.0\linewidth]{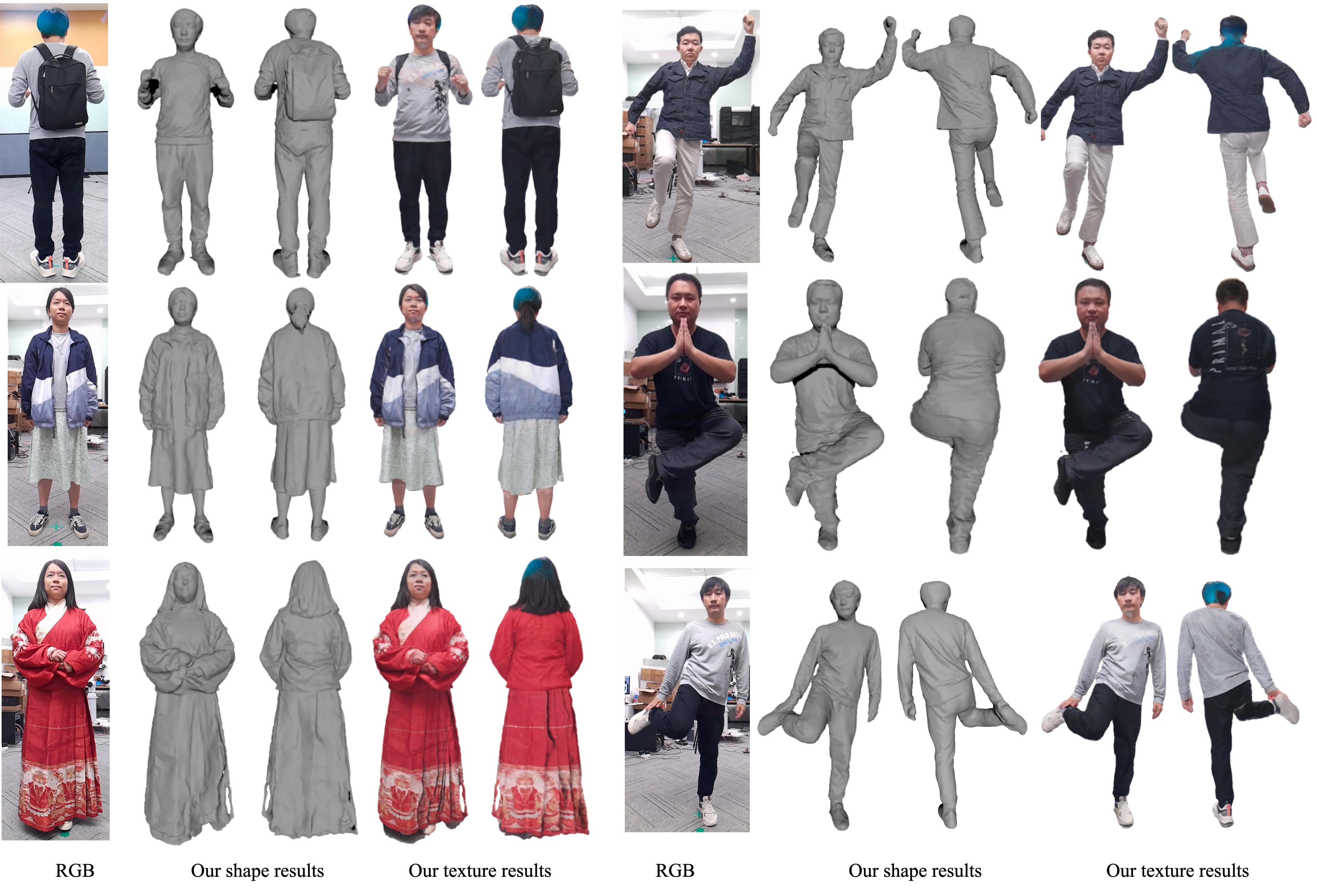}
\end{center}
\vspace{-0.15in}
    \caption{Results on real captured images. Our method generalizes well to various garments and poses and recovers high-quality shape details. }
\label{fig:real}
\vspace{-0.15in}
\end{figure*}

\subsection{Qualitative Results on Real Data}
We also experiment with our own real data, which is captured using 6 calibrated and synchronized Kinect cameras (only RGB images are used) surrounding the subject. We employ \cite{matting} to generate the image masks. Figure~\ref{fig:real} shows some of the results. As demonstrated  in these examples, our method does not rely on SMPL model estimation, enabling it to handle challenging cases with loose garments and unusual poses. All these examples are reconstructed using $6$ input images. Video results and additional  examples can be found in the supplementary files. 

\subsection{Ablation Study}
\begin{figure}
\begin{center}
    \includegraphics[width=1.0\linewidth]{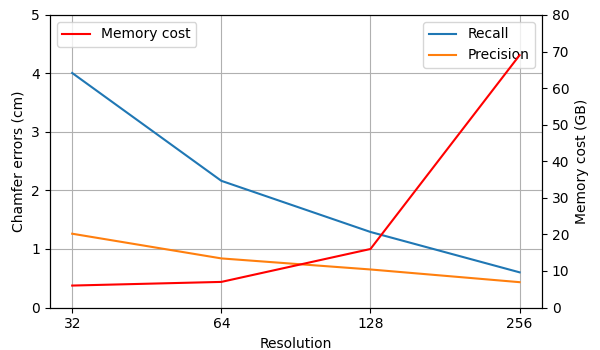}
\end{center}
\vspace{-0.15in}
    \caption{The performance and memory consumption with different  volume resolutions when using a na\"{\i}ve 3D UNet.}
\label{fig:dense-3dcnn}
\vspace{-0.15in}
\end{figure}

We conduct ablation studies to examine the effectiveness of our various design choices. To justify our system design, we also test a na\"{\i}ve implementation using a 3D UNet, which has the same architecture as the coarse stage with conventional 3D convolutions. We report the system performance and GPU memory consumption for different volume resolutions in Figure~\ref{fig:dense-3dcnn}. It is evident that higher volume resolution can significantly reduce shape errors, especially recall errors. However, GPU memory consumption also increases substantially, from 6G to 69G for volume resolution of 32 and 256, respectively. It is not feasible to scale this  na\"{\i}ve  implementation to a volume resolution of 512 on an A100.


\noindent \paragraph{\textbf{Ablation I: Shape ablation}}

To test the effectiveness of the normal feature and coarse level feature at the fine stage shape reconstruction, we conduct experiments using: AB1. only the coarse level feature; AB2. only the normal feature; and the proposed method, which uses both normal and coarse level features together.

We summarize the mean Chamfer and P2S errors of these different settings in Table~\ref{tab:Ablation_shape}. From AB1 and AB2, we can see that the normal features and the coarse level feature complement each other and should  both be included when computing the final TSDF.


\noindent \paragraph{\textbf{Ablation II: Texture ablation}}
To test the effectiveness of our texture estimation, we conduct experiments using: AB3. the network to directly compute a color field as in previous methods\cite{iccv2020PIFu,zheng2021deepmulticap}; AB4 \& AB5. after solving our blending weight field (with input images of 512 resolution by optimizing Equation~\ref{equation:color_loss}), we use 1K \& 2K images to compute the texture atlas map, respectively.



\begin{figure}
\begin{center}
    \includegraphics[width=1.0\linewidth]{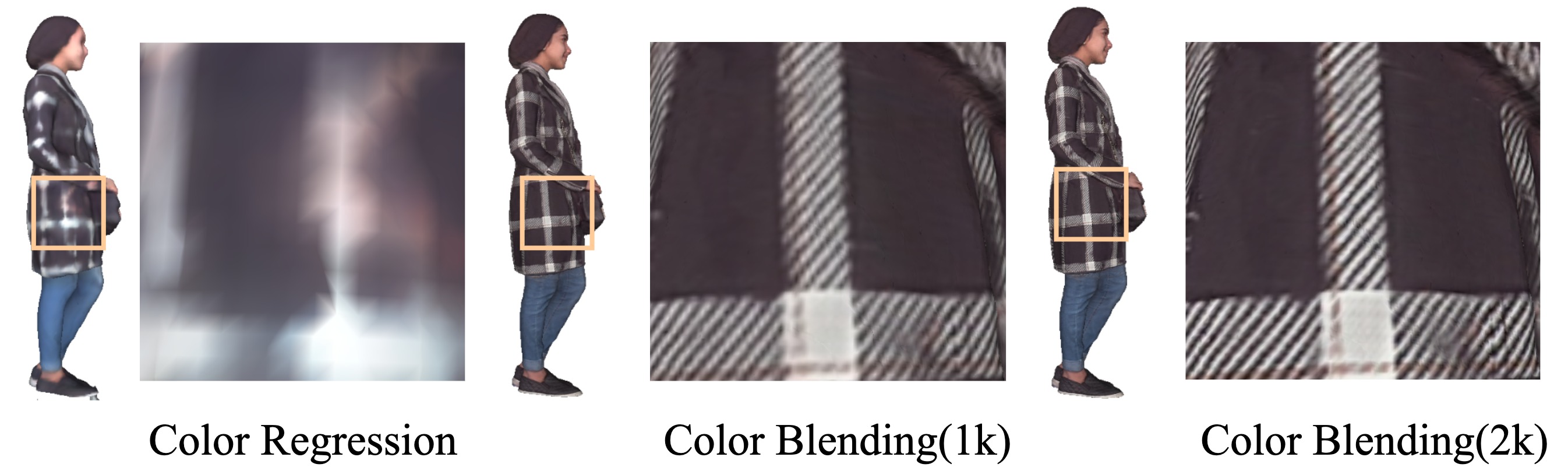}
\end{center}
\vspace{-0.15in}
    \caption{Results of different ablation settings for texture prediction. From left to right, they are the results of color field estimation, and our blending weight field estimation with 1K and 2K images, respectively.}
\label{fig:ablation_color}
\vspace{-0.15in}
\end{figure}

Table~\ref{tab:ablation_color} summarizes the PSNR and SSIM of the different settings. Firstly, AB3 has much poorer results than the other two settings  that involve blending weight estimation. It is evident that our blending weight strategy is crucial for generating sharp texture. Visualization in Figure~\ref{fig:ablation_color} reveals  that our strategy can produce sharp high-frequency texture details, while color field regression causes blurriness. 
AB4 and AB5 further demonstrate the scalability of our method. Both settings share the same blending weight volume $\mathbb{W}$, computed from input images of 512 resolution. We apply $\mathbb{W}$ to 1K and 2K images to compute the texture atlas map. Both settings generate high-quality results, while AB5 is slightly better. These two experiments demonstrate that our method can capture more texture details by re-evaluating the texture atlas map without the need for retraining.


\begin{table}
\begin{tabular}{c|cc|cc}
\hline
                  & \multicolumn{2}{c|}{\begin{tabular}[c]{@{}c@{}}Twindom \end{tabular}} & \multicolumn{2}{c}{\begin{tabular}[c]{@{}c@{}}MultiHuman\end{tabular}} \\
                  & PSNR                                       & SSIM                                        & PSNR                                  & SSIM                                   \\ \hline
AB3              &      23.776                                       &             0.854                             &             24.027                            &            0.860                          \\ \hline
AB4                &       26.309                                      &      0.862
&    26.033                                       &   0.866                                  \\ \hline
AB5                &        26.656                                      &     0.864
&     26.544                                       &   0.867                                   \\ \hline
\end{tabular}
\caption{Results of different ablation settings in texture prediction. Please refer to text for more details. 
}
\label{tab:ablation_color}
\vspace{-0.15in}
\end{table}



\begin{figure}
\begin{center}
    \includegraphics[width= 0.95\linewidth]{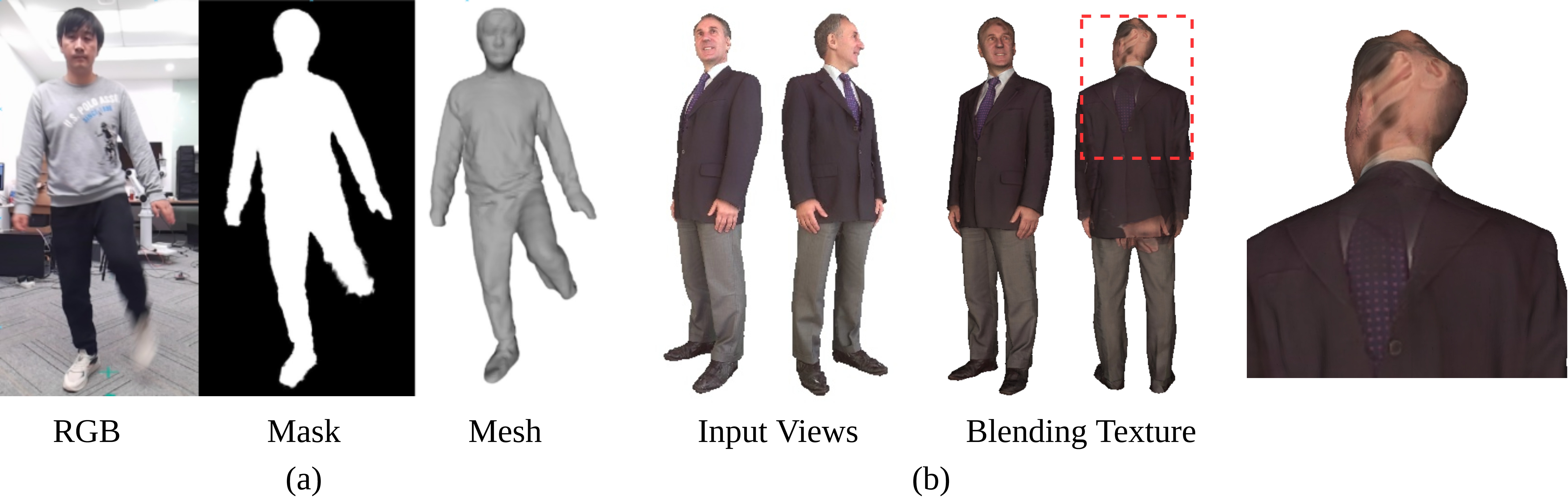}
\end{center}
\vspace{-0.15in}
    \caption{Failure cases of our method. (a) Due to the incomplete visual hull caused by poor matting, our method is unable to reconstruct the correct shape. (b) Unobserved regions have incorrect texture.}
\label{fig:failure_case}
\vspace{-0.15in}
\end{figure}

\subsection{Limitations and Future Work}
Our method has difficulties when addressing cases where the segmentation module fails. Figure~\ref{fig:failure_case} (a) shows such an example, where inaccurate segmentation due to motion blur results in an incomplete feature volume, consequently leading to poor final results. Inaccurate camera calibration may also contribute to poor feature volume construction and, subsequently, inferior shape results. As for texture prediction, our method computes texture maps by blending input images, which makes it difficult to handle unobserved regions, as shown in Figure~\ref{fig:failure_case} (b). In the future, we might consider employing an implicit function to address this problem.


\section{Conclusion}
We re-examine volumetric reconstruction for clothed humans and demonstrate that,  with proper system design, it can generate superior results than recent deep implicit methods. 
We find that a high volume resolution, such as 512 or above, effectively reduces the notorious  quantization error and capitalizes on the advantages of 3D CNNs for enhanced  exploration of local context information. 
To address the memory and computational challenges associated with high-resolution volumes, our method takes a coarse-to-fine approach, integrating sparse 3D CNN and voxel culling through  visual hulls and narrow bands.
Finally, it employs  an image-based rendering approach to compute the texture atlas map by blending input images with learned weights.
Extensive experiments demonstrate that our method significantly improves shape accuracy over SOTA techniques and captures vivid appearance details.

\begin{acks}
We would like to thank DGene company and Prof. Yebing Liu (Tsinghua University) for kindly providing human body datasets for our experiments.  
\end{acks}

\bibliographystyle{ACM-Reference-Format}

\end{document}